\pdfoutput=1

\documentclass{article}
\usepackage{ijcai23}

\usepackage{times}
\usepackage{soul}
\usepackage{url}
\usepackage[hidelinks]{hyperref}
\usepackage[utf8]{inputenc}
\usepackage[small]{caption}
\usepackage{graphicx}
\usepackage{amsmath}
\usepackage{amsthm}
\usepackage{booktabs}
\usepackage{algorithm}
\usepackage{algorithmic}
\usepackage[switch]{lineno}
\usepackage{tabularx}

\usepackage{algorithm}
\usepackage{algorithmic}

\usepackage{amsmath}
\usepackage{amsfonts}
\usepackage{xcolor}
\usepackage{amssymb}
\usepackage{amsthm}
\usepackage{mathtools}
\usepackage{comment}

\usepackage{physics}

\usepackage{subcaption}

\newcommand{\Lim}[1]{\raisebox{0.5ex}{\scalebox{0.8}{$\displaystyle \lim_{#1}\;$}}}

\makeatletter
\newtheorem*{rep@theorem}{\rep@title}
\newcommand{\newreptheorem}[2]{%
\newenvironment{rep#1}[1]{%
 \def\rep@title{#2 \ref{##1}}%
 \begin{rep@theorem}}%
 {\end{rep@theorem}}}
\makeatother

\newtheorem{theorem}{Theorem}
\newtheorem{prop}{Proposition}
\newtheorem{lemma}{Lemma}

\newtheorem{thm}{Theorem}
\newtheorem{proposition}{Proposition}
\newtheorem{lem}{Lemma}

\newtheorem{definition}{Definition}

\usepackage{newfloat}
\usepackage{listings}
\floatstyle{ruled}
\newfloat{listing}{tb}{lst}{}
\floatname{listing}{Listing}
\title{Towards a Better Understanding of Learning with Multiagent Teams}

\author{
David Radke\footnote{Contact Author}\and
Kate Larson\and
Tim Brecht\And
Kyle Tilbury\\
\affiliations
David R. Cheriton School of Computer Science, University of Waterloo\\
\emails
\{dtradke, kate.larson, brecht, ktilbury\}@uwaterloo.ca
}

\begin{document}

\newcommand*{\ourmean}[1]{\overline{#1}}

\newcommand{\oursubsub}[1]{\subsubsection{#1}}

\newcommand{\optional}[1]{\textcolor{YellowGreen}{#1}}

\newcommand{\todo}[1]{\textcolor{red}{\textbf{#1}}}

\newcommand{\done}[1]{\textcolor{orange}{\st{#1}}}

\newcommand{\addressreview}[1]{\textcolor{black}{#1}} 

\newcommand{\oldtext}[1]{\textcolor{cyan}{#1}}

\newcommand{\heading}[1]{\vspace{3pt}\noindent\textbf{#1 }}

\newcommand{\mysubsection}[1]{\vspace{3pt}\noindent\textbf{#1 }}

\newcommand{\camera}[1]{\textcolor{black}{#1}}

\newcommand{\newtext}[1]{\textcolor{black}{#1}} 

\newcommand{\vitem}{\vspace{-5pt}\item}

\newenvironment{guideline}{\vspace{0pt} \noindent \hrulefill \\ \emph{\bf \textcolor{blue}{GUIDELINE:}} \it }{\\ \vspace{-5pt} \hrule}

\newcommand{\squishbegin}{
 \begin{list}{$\bullet$}
  { \setlength{\itemsep}{0pt}
     \setlength{\parsep}{1pt}
     \setlength{\topsep}{1pt}
     \setlength{\partopsep}{0pt}
     \setlength{\leftmargin}{1.5em}
     \setlength{\labelwidth}{1em}
     \setlength{\labelsep}{0.5em} 
  } 
}

\newcommand{\squishtwobegin}{
 \begin{list}{$-$}
  { \setlength{\itemsep}{1pt}
     \setlength{\parsep}{1pt}
     \setlength{\topsep}{1pt}
     \setlength{\partopsep}{0pt}
     \setlength{\leftmargin}{1.5em}
     \setlength{\labelwidth}{1em}
     \setlength{\labelsep}{0.5em} 
  } 
}

\newcommand{\squishend}{
  \end{list}  
}

\newcommand{\experiment}{\vspace*{4pt}\noindent\textbf{Experiment Setup:\hspace{0.4em}}}
\newcommand{\experimentend}{}

\newcommand{\moveup}{\vspace{-8pt}}
\newcommand{\movecaptionup}{\vspace{-20pt}}
\newcommand{\movecaptionuptab}{\vspace{-17pt}}
\newcommand{\colfigwidth}{0.90\columnwidth}


\newcommand{\btable}[1]{\begin{table}[#1] \begin{center} }
\newcommand{\etable}[2]{\end{center} \vspace{-5pt} \caption{#2} \label{#1} \vspace{-15pt}\end{table}}

\newcommand{\wbtable}[1]{\begin{table*}[#1] \begin{center} }
\newcommand{\wetable}[2]{\end{center} \caption{#2} \label{#1} \end{table*}}

\newcommand{\xfigure}[5]{\begin{figure}[#1] \begin{center} \leavevmode \epsfxsize=#2 \epsfbox{#3} \end{center} \vspace{-12pt} \caption{#5} \label{#4} \end{figure}}

\newcommand{\xfigurewide}[5]{\begin{figure*}[#1] \moveup \begin{center} \leavevmode \epsfxsize=#2 \epsfbox{#3} \end{center} \movecaptionup \caption{#5} \label{#4} \end{figure*}}

\newcommand{\yfigure}[5]{\begin{figure}[#1] \begin{center} \leavevmode \epsfysize=#2 \epsfbox{#3} \end{center} \caption{#5} \label{#4} \end{figure}}

\newcommand{\xyfigure}[6]{\begin{figure}[#1] \begin{center} \leavevmode \epsfxsize=#2 \epsfysize=#3 \epsfbox{#4} \end{center} \caption{#6} \label{#5} \end{figure}}

\newcommand{\bfigure}[1]{\begin{figure}[#1]}
\newcommand{\efigure}[2]{\vspace{-8pt} \caption{#2} \label{#1} \end{figure}}

\maketitle

\begin{abstract}



While it has long been recognized that a team of individual learning agents can be greater than the sum of its parts, recent work has shown that larger teams are not necessarily more effective than smaller ones.
In this paper, we study why and under which conditions certain team structures promote effective learning for a population of individual learning agents.
We show that, depending on the environment, some team structures help agents learn to specialize into specific roles, resulting in more favorable global results.
However, large teams create credit assignment challenges that reduce coordination, leading to large teams performing poorly compared to smaller ones.
We support our conclusions with both theoretical analysis and empirical results.

\end{abstract}


\section{Introduction}
\label{sec:intro}


\newtext{
In Multiagent Systems (MAS), the study of how multiple agents work together in groups or teams has been a significant area of research for several decades~\cite{Pollack1986AMO,Radke2022Exploring}.
In settings with individual learning agents, teams are typically defined so that agents learn from their individual experiences but share rewards to varying extents~\cite{Baker2020EmergentTU,McKee2020SocialDA}.
In this paper, we investigate how teams and different team structures influence and guide the learning process of individual agents.
}

\newtext{
\camera{
We consider mixed-motive stochastic games, where agent interests are sometimes aligned and sometimes in conflict, often to varying extents~\cite{Dafoe2020OpenPI}.
Past work studying cooperation in mixed-motive domains typically assumes that a fully cooperative population (i.e., a single team) achieves the best results~\cite{yang2020learning,gemp2020d3c}.
}
However, recent work has indicated that non-fully cooperative populations can learn significantly more productive joint policies than a fully cooperative population~\cite{Durugkar2020BalancingIP,radke2022importance}. 
Even though a larger team has more agents at its disposal to perform tasks, smaller teams can achieve better global outcomes because agents learn more effective joint policies.
While this phenomenon has been observed across multiple domains, the underlying reasons have not been fully explored or understood.
}

\newtext{
This paper provides theoretical groundwork as to \emph{why}, and under \emph{which conditions}, smaller teams outperform larger teams.
We focus on two areas of how teams impact learning: 1) how the introduction of teammates initially improves the ability for individual agents to learn about valuable areas of the state space (Section~\ref{sec:teams_redistribute_reward}), and 2) 
how the credit-assignment problem (i.e. learning the value of taking a particular action) becomes more challenging as a team gets larger (Section~\ref{sec:negatives_team_size}).
}

\newtext{
Thus, this paper emphasizes the importance of teams and team structures to shape the learning problems and reward functions while agents learn: teams can help improve agents' learning processes to a point, but sub-optimal team structures can hinder learning.
This paper provides some theoretical underpinnings to help understand the impacts of teams on the learning processes of individual agents.
We make the following contributions:
}

\begin{itemize}
    \item We theoretically explore how teams can reduce the complexity of learning problems in certain environments.

    \item We show how sub-optimal team structures increase the difficulty for agents to identify valuable experiences, expanding previous work~\cite{arumugam2021information} to the multiagent team setting.

    \item \newtext{We validate our theory empirically with widely used multiagent testbeds.} 
\end{itemize}

\section{Related Work}
\label{sec:related_work}





Research on multiagent teams with individual learners includes frameworks where agents share mental models or plans~\cite{Pollack1986AMO,Pollack90plansas,Tambe1997TowardsFT}, ad hoc teamwork~\cite{Stone2010AdHA}, and multiagent reinforcement learning (MARL) teams with shared rewards~\cite{Radke2022Exploring,radke2022importance}.
In MARL with individual learners, algorithms to improve decentralized learning from a shared team reward have defined multiple learning rates~\cite{matignon2007hysteretic} or regulated replay buffers~\cite{palmer2018negative}.
While the behavior of reward-sharing agents is often studied, the underlying impact on the learning process and environment, analyzed in this work, are often overlooked.

\addressreview{
Groups sharing a team reward can cause a credit assignment problem -- where it may be difficult to determine the value of actions on an objective (i.e., reward for specific actions).
}
\newtext{
In cooperative settings with control over all agents, an existing approach to reduce credit assignment issues and achieve effective teamwork is the centralized training decentralized execution (CTDE) methodology, where value decomposition algorithms are effective~\cite{Rashid2018QMIXMV}.
During a centralized training phase, value decomposition learns to distribute team (or sub-team~\cite{phan2021vast}) reward among it's members to guide learning towards favorable policies.
This methodology relies on a fully cooperative population, access to agents' internal learning algorithms, and a separate training phase.
}
\camera{
Our work focuses on mixed-motive environments with individual learners where these assumptions do not hold and a cooperative population is often assumed to achieve the best result~\cite{yang2020learning,dong2021birds}.
}

\newtext{
In single agent settings, ideas like reward shaping have been well studied.
}
Reward shaping adds extra rewards to help agents discover valuable state-action pairs; however, this typically follows a fixed reward redistribution policy in stationary environments~\cite{ng1999policy,wiewiora2003principled}.
RUDDER~\cite{arjona2019rudder} uses supervised learning as a way to dynamically redistribute reward to reward-causing state-action pairs in an agent's memory.
Options~\cite{sutton1999between,konidaris2007building} and Hierarchical RL (HRL)~\cite{nachum2018data,gurtler2021hierarchical} create artificial sub-goals in the environment (instead of in memory) to guide an action policy towards potentially valuable state-action pairs as it learns.
In curriculum learning, agents train with increasingly complex tasks to discover valuable state-action pairs~\cite{wang2021survey}.

Many of these single-agent approaches do not directly translate to the nonstationary domains of MARL.
Our work demonstrates how, under certain conditions, teams have emergent features that influence the learning process similarly to the aforementioned single-agent frameworks.
We show how teammates redistribute reward to valuable reward-causing state-action pairs.
This rate of redistribution is not fixed, but rather the reward signal increases in strength as agents learn to act and obtain more reward (similar to RUDDER and curriculum learning).
We draw comparisons with these ideas to highlight how research with teams has ties to several fundamental problems across RL.


\section{Background}
\label{sec:background}


A typical environment in which multiagent teams operate is a stochastic game: a generalized Markov Decision Process (MDP) with multiple agents~\cite{shapley1953stochastic}.
We model our base environment as a stochastic game $\mathcal{G}=\langle \mathcal{N}, \{S\}_{i\in N}, \\ \{A\}_{i\in N}, \{R\}_{i\in N}, P, \gamma, \Sigma \rangle.$
$\mathcal{N}$ is our set of agents with size $N \in \mathbb{N}$ that learn online from experience and $S$ the state space, observable by all agents.
We use $s_i \in S_i$ to define a single state observed by agent $i$ in the set of states $i$ observes.
$A = A_1\times \ldots \times A_N$ is the joint action space for all agents where $A_{i}$ is the action space of agent $i$ and $\mathbf{a}\in A$.
$R = R_1 \times \ldots \times R_N$ is the joint reward space for all agents where $R_{i}$ is the reward function of agent $i$, defined as $R_i: S \times A \times S \mapsto \mathbb{R}$, a real-numbered reward for taking an action in a state and transitioning to the next state.
\camera{We assume all agents have identical (deterministic) reward functions (i.e., receive the same reward for the same behavior).}
The transition function, $P:S \times A\mapsto \Delta(S)$, maps a joint state and joint action into a next joint state with some probability and $\gamma$ is the discount factor.
$\Sigma$ represents the policy space of all agents and the policy of agent $i$ is $\pi_i:S \mapsto A_{i}$, which specifies an action that agent $i$ should take in an observed state.\footnote{We can also allow for randomized policies.}
\camera{
$\boldsymbol\pi_N = \pi_1\times \ldots \times \pi_N$ denotes the joint policy of $N$ agents.
}

We assume agents are independent learners and learn policies $\pi_i$ based on their individual observations $s_i$ and actions $a_i$, but the rewards that they receive can depend on the actions of others.
Bold notation represents joint states $\vb{s}^t$ and joint actions $\vb{a}^t$ of all agents.
In particular, at each timestep, $t$, agent $i$ receives reward $R_{i}^t(\vb{s}^t,\vb{a}^t,\vb{s}^{t+1})$.
Learning agents seek to learn policies which maximize their sum of discounted future rewards, $V_i=\sum_{t=0}^\infty \gamma^t R_{i}^t(\vb{s}^t,\vb{a}^t,\vb{s}^{t+1})$.

\subsection{Model of Teams}
\label{sec:teams_model}


We define a team as a subset of individual learning agents that share common interest for team-level goals through a shared reward.
Given a population, multiple teams with different preferences and interests may co-exist~\cite{Mathieu2001MTSs}; real-world examples include work teams within the same company, position teams that make up a sports team, or faculties within a university.
\emph{Team structure} denotes the number and size of teams in a population.
Our theory depends on all $N$ agents and scales to environments with multiple teams; thus, we use ``team structure'' to denote ``team size'' when considering a single team to remain general.
We refer to the set of all teams as $\mathcal{T}$, the set of teams agent $i$ belongs to as $\mathcal{T}_i$, and a specific team that agent $i$ belongs to as $T_{i} \in \mathcal{T}_i$.
For readability, we define the size of a team $|T_i| = n$.

Agents on a team continue to observe only their individual state $s_i$ and take individual actions $a_i$ but share rewards through a shared team reward function $TR_{i[n]}$,  isolating the team infrastructure to only the reward signal.
Replacing $R_{i}$ in agents' optimization problems' with $TR_{i[n]}$, agents learn to maximize the sum of discounted future \textbf{team} rewards.
Any deterministic function for reward-sharing can define $TR_{i[n]}$ \camera{as long as every agent in a team gets some amount of the team's reward.}
We use:

\begin{equation}
    TR_{i[n]} = \frac{\sum_{j\in T_i} R_j(S,A,S)}{n},
    \label{eq:team_reward}
\end{equation}

\noindent
where teammates share their rewards equally to be consistent with past work~\cite{Baker2020EmergentTU,Radke2022Exploring}.



Much of our theory relies on features of agent or team trajectories.
We define $\tau_i = \{(s_i^1, a_i^1), (s_i^2, a_i^2), \cdots, (s_i^{H}, a_i^{H})\}$ to be a trajectory of individual state-action pairs generated by agent $i$ following $\pi_i$ over $H$ timesteps.
A joint policy for $T_i$ is the collection of individual behavior policies of all $n$ agents in $T_i$, $\boldsymbol\pi_{T_i}$.
A joint trajectory for team $T_i$, $\boldsymbol\tau_{T_i}$, is the collection of joint state-action pairs generated by agents in $T_i$.
We are required to index trajectories in three ways: first $\boldsymbol\tau_{T_i}^{t} = (\vb{s}_{T_i}^t, \vb{a}_{T_i}^t)$, second $\boldsymbol\tau_{T_i}^{1:t-1} = \{(\vb{s}_{T_i}^1, \vb{a}_{T_i}^1), \cdots, (\vb{s}_{T_i}^{t-1}, \vb{a}_{T_i}^{t-1})\}$, and third $\boldsymbol\tau_{T_i}^{-t}$ is the $H$-timestep trajectory without timestep $t$.
Let
$Z(\boldsymbol\tau_{T_i})$ be a random variable denoting the team return obtained after team $T_i$ completes the joint trajectory $\boldsymbol\tau_{T_i}$ following their individual policies that compose $\boldsymbol\pi_{T_i}$.
We define $Z_{T_i} \triangleq Z(\vb{s}_{T_i},\vb{a}_{T_i})$ to be a random variable denoting the team reward observed at the joint state of all teammates $\vb{s}_{T_i}$ having taken joint action $\vb{a}_{T_i}$ and following their individual policies thereafter.
Note that $\vb{s}_{T_i}$ is dependent on all $N$ agents in the system by definition of stochastic games.


\section{Identifying Valuable State-Action Pairs}
\label{sec:teams_redistribute_reward}





We study the most restrictive case where teammates have no communication or coordination mechanisms and focus only on features of the team reward function.
By the definition of a stochastic game, rewards obtained from the environment depend on the joint states and actions of all agents.
\newtext{
Thus, there can exist \emph{reward-causing state-action pairs} -- experiences that may not yield reward themselves, but allow reward to be obtained elsewhere in the environment~\cite{arjona2019rudder}.
}
\newtext{
Identifying these pairs can be challenging, since independently each state may provide little or no reward.
}

\newtext{
We want to understand when teams of agents can leverage these reward-causing state-action pairs.
We distinguish between the direct reward an agent receives from the environment when transitioning into their own observed state $s_i^{t+1}$, $R_{i}^t(\vb{s}^t,\vb{a}^t,s_i^{t+1})$, and the team reward, $TR_{i[n]}$.
We identify an environmental property where the team reward signal is stronger than the individual reward signal (i.e., $\mathbb{E}\left[ R_{i}^t(\vb{s}^t,\vb{a}^t,s_i^{t+1}) \right] < \mathbb{E} \left[ TR_{i[n]} \right]$) because of the reward-causing state-action pair effect.
}
\addressreview{
This signal causes agents to become more attracted to, and thus learn to execute, reward-causing state-action pairs more often.
}


\begin{figure}[t]
    \centering
    \includegraphics[width=\linewidth]{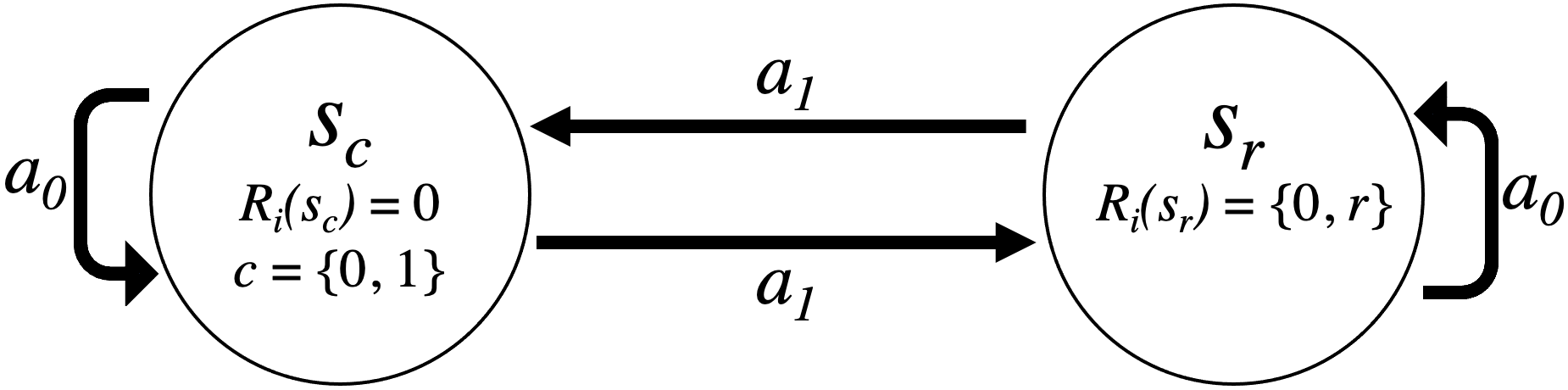}
    \caption{Diagram of our two state example environment.}
    \label{fig:2_states}
\end{figure}

Consider the two state environment shown in Figure~\ref{fig:2_states}.
This environment can support any number of agents ($N \geq 1$), and the state transitions and rewards depend on the joint action of all $N$ agents.
For simplicity, we assume the existence of only one team (i.e., $n = N$).
A stochastic game ($n=2$) diagram is given in Figure 5 in Appendix A.
There exists two physical states that agents individually observe: $s_c$ and $s_r$.
Agents have two actions: stay at their current state ($a_0$) or move to the other state ($a_1$).
The ``$c$'' in $s_c$ corresponds to a binary signal (explained below) and ``$r$'' in $s_r$ refers to a reward state.

There is never an \emph{environmental} reward given to agent $i$ for transitioning to $s_c$, thus $R_{i}^t(\vb{s}^t,\vb{a}^t,s_c^{t+1}) = 0$.
However, any agent (regardless of team affiliation) visiting $s_c$ changes a binary signal $c$ that allows reward to be collected at $s_r$.
Thus, the possible rewards (dependent on $c$) given to any agent in $s_r$ are $R_{i}^t(\vb{s}^t,\vb{a}^t,s_r^{t+1}) = \{0, r\}$, where $r > 0$.
When agent $i$ individually transitions to $s_r$, their reward (before sharing with their team) is $R_{i}^t(\vb{s}^t,\vb{a}^t,s_r^{t+1}) = 0$ if $c=0$, and their reward is $R_{i}^t(\vb{s}^t,\vb{a}^t,s_r^{t+1}) = r$ if $c=1$.
\newtext{
Once reward is consumed at $s_r$, $c$ has to be reset by visiting $s_c$ again.
}
Thus, the reward-causing state-action pair in this environment is to ``visit $s_c$'', causing a reward to be obtained when visiting $s_r$.
With teams, the rewards given to individual agents are transformed into the team reward by Equation~\ref{eq:team_reward} for agents to learn.

We can generalize features of this environment to support theory about how teams impact learning under certain conditions.
In doing so, our theory is applicable to any multiagent environment where the following assumptions hold:



\begin{enumerate}
    \item Agents' policies are initialized at random and fully explore the state space (in the limit).
    \item The environment yields mid-episode rewards (not only at termination state) and any agent can collect rewards.
    \item Executing a reward-causing state-action pair returns a minimum reward in the environment if the agent is not in a team (e.g., visiting $s_c$ returns a reward of 0, the minimum reward of the environment in Figure~\ref{fig:2_states}).
\end{enumerate}

\begin{theorem}
There exists an environment where increasing team size increases the probability of an agent receiving a reward for executing any reward-causing state-action pair that is greater than if they were not in a team.
\label{thm:reward_increase}
\end{theorem}

\newtext{
The proof of this theorem is in Appendix B.
This theorem has direct implications on the policy that $i$ learns -- more positive reward for executing a particular state-action pair will cause $i$ to execute that pair more often.
A larger team monotonically increases the \textbf{probability} that any teammate will be in a corresponding reward state and instantaneously share this reward with $i$ through $TR_{i[n]}$.
From the perspective of individual agents, this distributes the environment's reward function to other areas of the state space.
}


\newtext{
Consider our two state environment in Figure~\ref{fig:2_states}.
Agents individually receive a reward of 0 for visiting $s_c$ but receive a reward of $r$ for visiting $s_r$ when $c=1$.
Without teammates ($n=1$), agent $i$ only receives the environmental reward when visiting $s_c$ (i.e., reward of 0).
\emph{With} teammates, $i$ receives a reward (through $TR_{i[n]}$) of at least $\frac{r}{n} > 0$ when visiting $s_c$ if at least one teammate is visiting $s_r$.
This probability increases if there are more teammates.
The implications of this is that $i$ will learn the benefit of executing it's part in a reward-causing state-action pair, leading it to execute this role more often.
}

\section{Teams Impact on Credit Assignment}
\label{sec:negatives_team_size}









Whereas the previous section showed how teams increase the reward agents receive for executing reward-causing state-action pairs (i.e., visiting $s_c$), we now analyze the relationship between team structure and the distribution of rewards across all state-action pairs.
In this section, we use information theory to explore how sub-optimal team structures impact the ability for agents to perform credit assignment.

\subsection{Information Sparsity in Single-Agent Settings} 
\label{sec:info_in_sgs}

Credit assignment is concerned with identifying the value past actions on observed future outcomes and rewards.
In single agent Markov Decision Processes (MDPs), information theory has been used to formalize conditions which make credit assignment infeasible, such as when the environment does not provide enough information (through reward) for an agent to learn~\cite{arumugam2021information}.
We expand this concept to our setting.
Let $s_i \in S_i$ and $a_i \in A_i$ represent any arbitrary state and action by an agent $i$ within their individual state and action spaces.
Following the single-agent case definitions in~\cite{arumugam2021information} (i.e., if $N=1$), let $Z_i$ be a random variable denoting the return for a single agent having taken action $a_i$ in state $s_i$, and following $\pi_i$ thereafter.
The information gained by $\pi_i$ is a random variable, defined as:

\begin{equation}
    \mathcal{I}^{\pi_{i}}_{s_i,a_i} = D_{KL} ( p(Z_i|s_i,a_i) || p(Z_i|s_i)),
    \label{eq:info_adv}
\end{equation}

\noindent
the Kullback-Leibler (KL) divergence between $p(Z_i|s_i,a_i)$ and $p(Z_i|s_i) = \Sigma_{a_i \in A_i} \pi_{i}(a_i|s_i) p(Z_i|s_i,a_i)$ (the distribution over returns for random state-action pairs conditioned on a particular state and action (i.e., the $Q$-value) versus only a particular state $s_i$ (i.e., the value function)).
Equation~\ref{eq:info_adv} is the distributional analogue to the advantage function in RL, $A^{\pi_{i}} = Q^{\pi_{i}}(s_i,a_i) - V^{\pi_{i}}(s_i)$, the difference between the value of taking action $a_i$ at state $s_i$ and the expected value of state $s_i$.
Let $d^{\pi_{i}}$ be the distribution of states visited and actions taken by $i$'s policy.
The expected amount of information carried by the actions of $\pi_{i}$ about the return of those state-action pairs is defined as:

\begin{equation}
    \mathcal{I}(A_i;Z_i|S_i) = \mathbb{E}_{(s_i,a_i)\sim d^{\pi_{i}}} \left[ D_{KL} ( p(Z_i|s_i,a_i) || p(Z_i|s_i)) \right].
    \label{eq:exp_info}
\end{equation}

\noindent
Difficulties with credit assignment emerge when $\mathcal{I}(A_i;Z_i|S_i)$ is small enough that the actions of a policy carry almost no correlation with the reward signal.
Prior work defined an $\epsilon$-information sparse MDP as $\mathcal{I}^{\pi_{i}}(A_i;Z_i|S_i) \leq \epsilon$ for any initial policy at the beginning of training~\cite{arumugam2021information}.
However, Equation~\ref{eq:exp_info} and $\epsilon$-information sparsity do not fully translate to the multiagent team setting since they only consider the \emph{expected} information.
Teams modify the distribution, or variance, of information (Equation~\ref{eq:info_adv}) across state-action pairs conditioned on the experienced values of the team reward random variable, $Z_{T_i}$.

For example, consider a \textbf{non}-$\epsilon$-information sparse single-agent MDP environment where one state-action pair yields reward $r$ and every other state-action pair gives a reward of zero.
If $r$ is divided evenly and distributed so that every state-action pair yields the same reward, $\mathcal{I}(A_i;Z_i|S_i)$ is unchanged (due to expectation) but the agent's policy carries no correlation with the reward signal.
The agent would now be unable to learn the same optimal policy as before (i.e., visiting the state which previously yielded $r$).

\subsection{Information Sparsity with Teams}

We enrich the definition of information sparsity in the context of stochastic games.
This must consider two aspects of information.
First, the expected information gained by $i$'s individual policy given their team reward function, substituting $Z_i$ with $Z_{T_i}$ in Equation~\ref{eq:info_adv}, $\mathcal{I}^{\pi_{i}}(A_i;Z_{T_i}|S_i)$.
Second, we must also consider the variance of information gained by $i$'s policy over the distribution of their individual state-action pairs given their team reward function,
var$\left[ \mathcal{I}^{\pi_{i}}_{S_i,A_i} (Z_{T_i}) \right]$ (see Appendix D for the extended KL-Divergence derivation).

\begin{definition}
Given a stochastic game with non-stationary policy class $\pi_{H}$, let $\boldsymbol\pi_N^0$ denote the set of initial policies for all $N$ agents employed at the very beginning of learning.
For small constants $\epsilon > 0$ and $\mu > 0$, a stochastic game is ($\epsilon,\mu$)-information sparse if:
\[ \sup_{\pi_i \in \boldsymbol\pi_N^0} \mathcal{I}^{\pi_{i}} (A_i;Z_{T_i}|S_i) \leq \epsilon, \]
or
\[ \sup_{\pi_i \in \boldsymbol\pi_N^0} \textrm{var} \left[ \mathcal{I}^{\pi_{i}}_{S_i,A_i} (Z_{T_i}) \right] \leq \mu. \]
\label{def:eps_mu_sparse}
\end{definition}

Definition~\ref{def:eps_mu_sparse} states that the actions of any agent $i$'s policy given their shared team reward function must carry enough information with high enough variance for $i$ to be able to learn.
Otherwise, the stochastic game is considered ($\epsilon,\mu)$-information sparse.
Low variance of information is detrimental to credit assignment since an agent would receive similar rewards regardless of their policy.
By redistributing rewards, teams can significantly modify $\textrm{var} \left[ \mathcal{I}^{\pi_{i}}_{S_i,A_i} (Z_{T_i}) \right]$ compared to settings without teams.

\subsection{Risks of Sub-Optimal Team Structure}
\label{sec:decrease_state_info}

We now analyze convergence properties of the team reward as a function of team size, conditioned on the behavior of all $N$ agents (i.e., global team structure).
Since $TR_{i[n]}$ is determined by the experiences of all teammates, we focus on the joint policy of agents in $T_i$, $\boldsymbol\pi_{T_i}$, which determines the team return over a joint trajectory, $Z(\boldsymbol\tau_{T_i})$.

First, assume we have a (\textbf{non}-($\epsilon,\mu$)-information sparse) stochastic game with $N$ agents and no teams.
By Definition~\ref{def:eps_mu_sparse}, this environment has enough information with high enough variance for individual agents to be able to learn.
Creating teams of agents in this game impacts the team structure and the reward signals agents learn from.
In the previous section, we showed how increasing a team's size increases the probability of $i$ receiving a better reward signal for executing a reward-causing state-action pair than without teams.
However, effectively identifying these pairs depends on an appropriate team structure.

Next, we theoretically show how a sub-optimal team structure can transform this non-($\epsilon,\mu$)-information sparse stochastic game into an ($\epsilon,\mu$)-information sparse stochastic game by decreasing the variance of information through $TR_{i[n]}$ below $\mu$ as team size increases in the limit.
In practice, $N$ (or the size of a team $n$) need only be sufficiently large to reduce the variance of information below $\mu$ as agents are grouped together in a team.
This has implications on an agent's ability to perform credit assignment and learn an effective policy.
To formalize this, we leverage a finding in~\cite{arumugam2021information} which we adapt to the multiagent team setting which equates information with reward entropy.
The proof is in Appendix C.

\begin{prop}
Let $\boldsymbol\pi_{T_i}$ be the joint fixed behavior policy of agents in $T_i$ that generates a joint trajectory of experiences $\boldsymbol\tau_{T_i}$, where the randomness of state-action pairs in $\boldsymbol\tau_{T_i}$ depends on all $N$ agents (by the definition of a stochastic game).
Let $TR_{i[n]}^t$ be a random variable denoting the team reward at any timestep $t$ (where the randomness of the deterministic reward follows from the randomness of the joint state-action pairs of individual agents in $T_i$ at time $t$, depending on all $N$ agents, $\boldsymbol\tau_{T_i}^{t}$).
It follows that:
\[ \mathcal{I}(Z(\boldsymbol\tau_{T_i}); \boldsymbol\tau_{T_i}^{t} | \boldsymbol\tau_{T_i}^{-t}) = \mathcal{H}(TR_{i[n]}^{t} | \boldsymbol\tau_{T_i}^{1:t-1}). \]
\label{prop:info_entropy}
\end{prop}

The equality states that the information of the joint policy for team $T_i$ at time $t$ is equal to the entropy, a measure of missing information or uncertainty~\cite{shannon1948mathematical}, of the team reward at timestep $t$, $TR_{i[n]}^t$, given the team-wide joint trajectory up to time $t$.
For example, if $TR_{i[n]}^t$ returns the same value at each timestep regardless of the joint policy, the entropy of this reward function is zero and the information gained by the team's joint policy, and each agent's individual policy, is zero.


We next show how the variance of $TR_{i[n]}$ converges to zero as a function of increasing team size.
The variance describes the distribution of potential team rewards given the randomness of state-action pairs experienced by agents in $T_i$.
The proof is in Appendix C.

\begin{lemma}
The team reward random variable $TR_{i[n]}$ for any state-action pair converges to the mean environmental reward (mean of any agent's individual reward function) as team size increases in the limit (i.e., $TR_{i[n]}(\vb{s}^t, \vb{a}^t, \vb{s}^{t+1}) \rightarrow \overline{R_i}$ as $n \rightarrow \infty$).
\label{lem:reward_distribution_exp}
\end{lemma}


Using Proposition~\ref{prop:info_entropy} and Lemma~\ref{lem:reward_distribution_exp}, we conclude that the information carried by the joint policy of teammates over the joint trajectory $\boldsymbol\tau_{T_i}$ converges to zero as team size increases.
The proof is in Appendix C.

\begin{theorem}
The information in a stochastic game at time $t$, $\mathcal{I}(Z(\boldsymbol\tau_{i}); \boldsymbol\tau_{i}^{t} | \boldsymbol\tau_{i}^{-t})$, converges to 0 as the size of a team, $n$, increases in the limit.
\label{thm:entropy_to_zero}
\end{theorem}

\newtext{
Since $\mu > 0$, defining larger teams \textbf{will} make a non-($\epsilon,\mu$)-information sparse stochastic game into an ($\epsilon,\mu$)-information-sparse stochastic game if the team is too large.}
In this setting, $TR_{i[n]}$ would not provide enough information about agents' individual policies and has implications on credit assignment, leaving agents unable to learn.

Theorems~\ref{thm:reward_increase} and \ref{thm:entropy_to_zero} imply the existence of an \emph{optimal} team structure.
Increasing the size of teams can help agents identify reward-causing state-action pairs (Theorem~\ref{thm:reward_increase}); however, sub-optimal team structure carries the risk of infeasible credit assignment (Theorem~\ref{thm:entropy_to_zero}).
\addressreview{
Since $\epsilon$ and $\mu$ are domain dependent, discovering the best team structure to help agents learn remains subject to many domain specific variables.
We can theoretically define a general rule that this team structure follows: $\max n$ s.t. $\sup_{\pi_i \in \boldsymbol\pi_N^0} \mathcal{I}^{\pi_{i}} (A_i;Z_{T_i}|S_i) > \epsilon$ and $\sup_{\pi_i \in \boldsymbol\pi_N^0} \textrm{var} \left[ \mathcal{I}^{\pi_{i}}_{S_i,A_i} (Z_{T_i}) \right] > \mu$.
To investigate features of this optimal structure in practice, we next empirically evaluate teams across multiple multiagent domains that support increasingly large populations of agents.
}



\section{Experimental Environments}
\label{sec:evaluation}




\newtext{
We study the impact of team structure on learning in four environments: 4-States, an Iterated Prisoner's Dilemma (IPD)~\cite{Rapoport1974PrisonersD}, Cleanup Gridworld Game~\cite{SSDOpenSource}, \newtext{and Neural MMO (NMMO)~\cite{suarez2019neural}}.
We specifically choose environments that support any number of agents (there is no fixed maximum), are not zero-sum, and have a diverse set of reward-causing state-action pairs.}
Our evaluation is designed to study various levels of RL algorithm and environmental complexity.
\newtext{
Due to space, the IPD details and results, including an analysis of policy network smoothness (influence of $n$ on the maximum eigenvector of the policy Hessian matrix, $\lambda_{\max}$), appear in Appendix F.
We give preliminary overviews of the environments here and include all details of environments in Appendices E, F, G, and H.
}

\paragraph{4-States.}
4-States is a simple, partially observable stochastic game based on the two state environment in Figure~\ref{fig:2_states} \addressreview{(Figure 6
in Appendix E.1)}.
In addition to $s_r$ and $s_c$, we add two new states, $s_3$ and $s_4$, which generate no reward.
Agents simultaneously choose among four actions: stay still or move to any of the other three states.
An action transitions agents to their intended next state with 90\% probability and to another random state with $10\%$ probability.
We fix $|\mathcal{T}| = 1$ and increase $n$ by a factor of 2 to remove the impact of other teams on the binary signal.
Agents use Tabular $Q$-Learning~\cite{sutton2018reinforcement} with $\gamma = 0.9$ and $\epsilon$-exploration ($\epsilon=0.3$) for 50 trials of 1,000 episodes (100 steps each).
\paragraph{Cleanup Gridworld Game.}
Cleanup~\cite{SSDOpenSource} is a temporally and spatially extended Markov game representing a sequential social dilemma.
We keep the underlying environment unchanged from previous setups~\cite{Leibo2017MultiagentRL} with the exception of the team reward.
Agent observability is limited to an egocentric 15 $\times$ 15 pixel window and
collecting an apple yields +1 reward (apple regrowth rate is dependent on the cleanliness of an adjacent river).
We set $|\mathcal{T}|=1$ and increase team size to remove impacts of other teams on the conditional reward structure.
We implement Proximal Policy Optimization (PPO)~\cite{PPO2017} agents for 10 trials of $1.6 \times 10^8$ environmental steps (1,000 timesteps per-episode) using the Rllib RL library.\footnote{\url{https://docs.ray.io/en/latest/rllib/index.html}}

\begin{figure*}[t]
    \centering
    \begin{subfigure}[b]{0.31\linewidth}
        \centering
        \includegraphics[width=\linewidth]{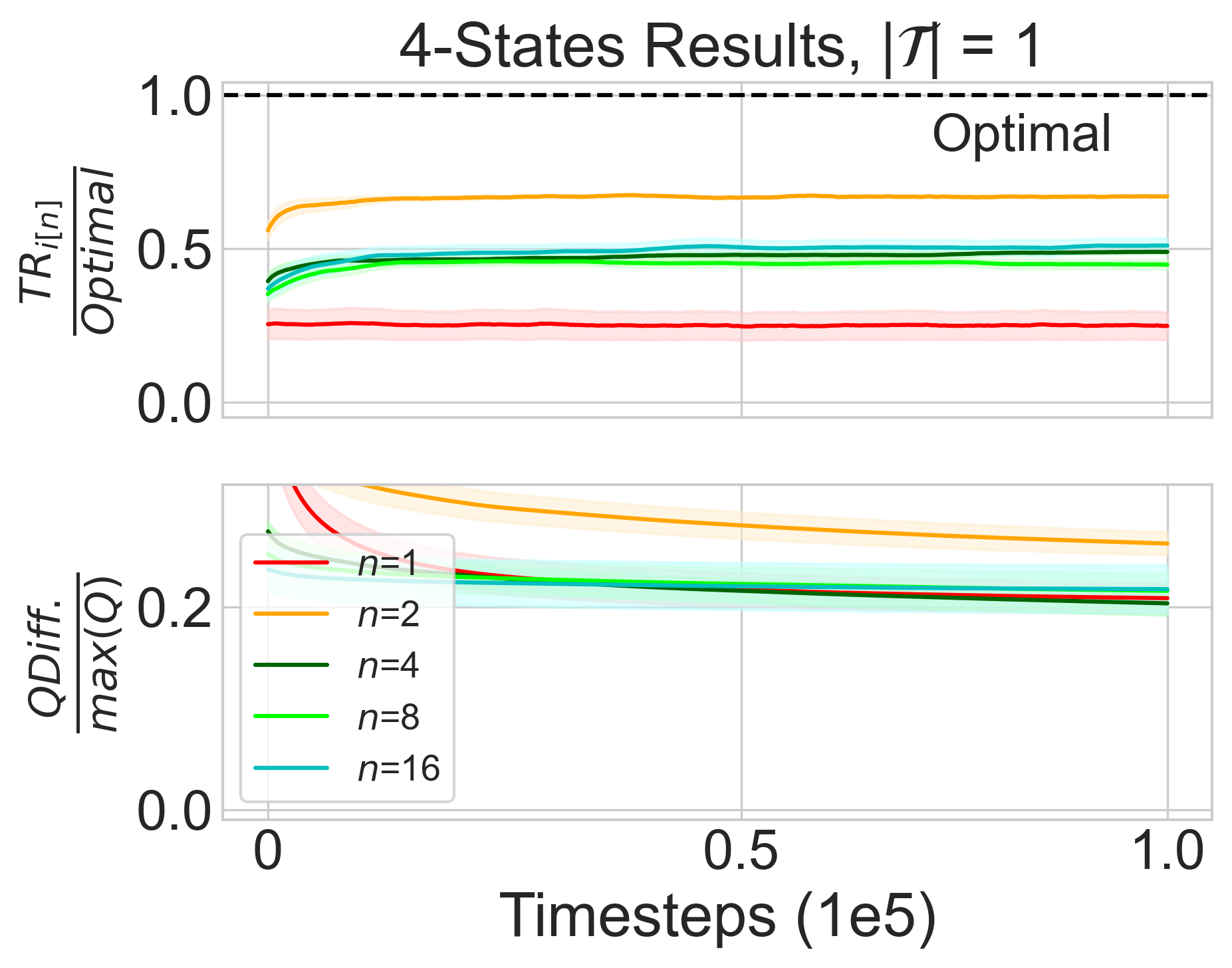}
        \caption{4-States Results.}
        \label{fig:4_state_results}
    \end{subfigure}
    \begin{subfigure}[b]{0.315\linewidth}
        \includegraphics[width=\linewidth]{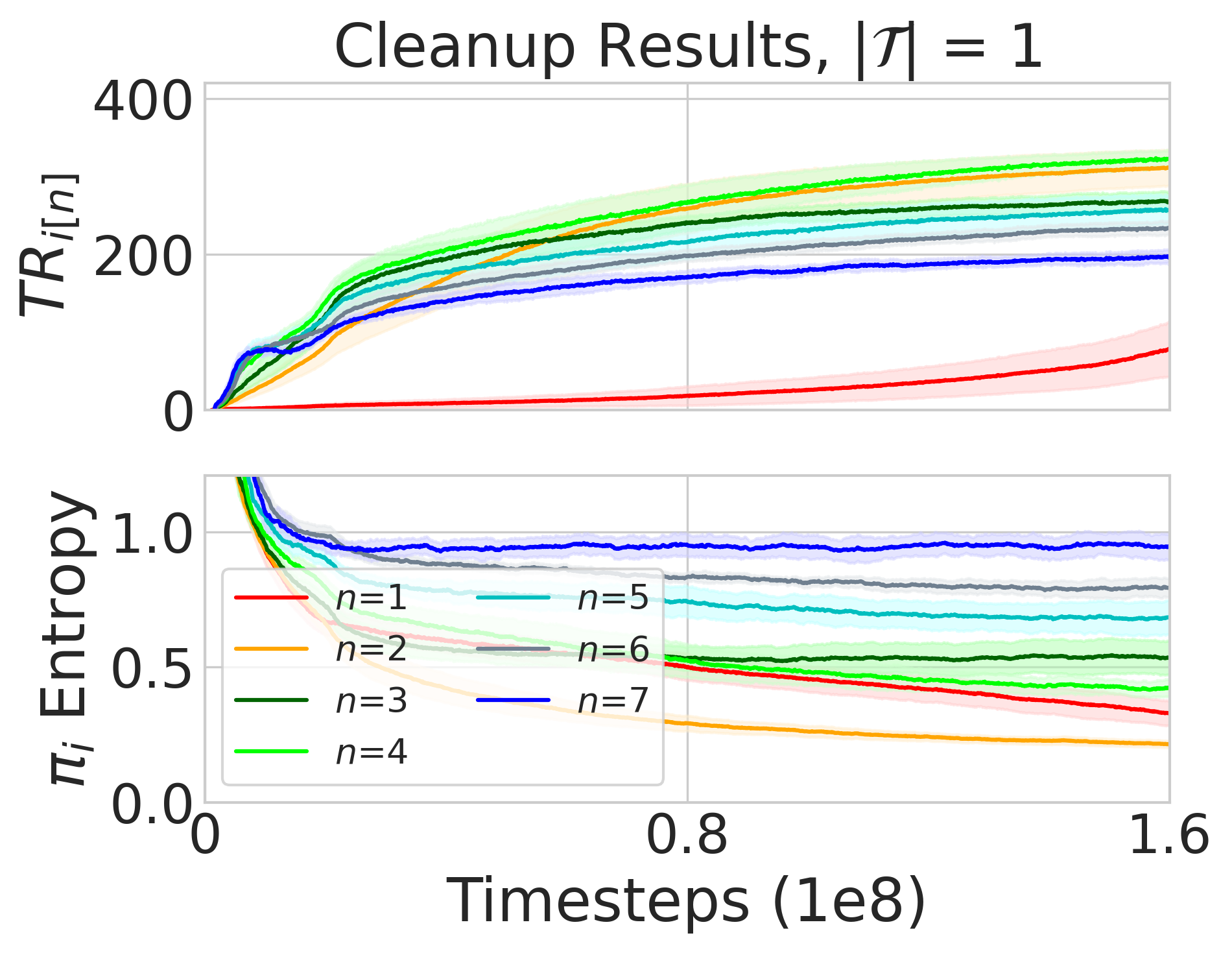}
        \caption{Cleanup Results.}
        \label{fig:cleanup_results}
    \end{subfigure}
    \begin{subfigure}[b]{0.305\linewidth}
        \includegraphics[width=\linewidth]{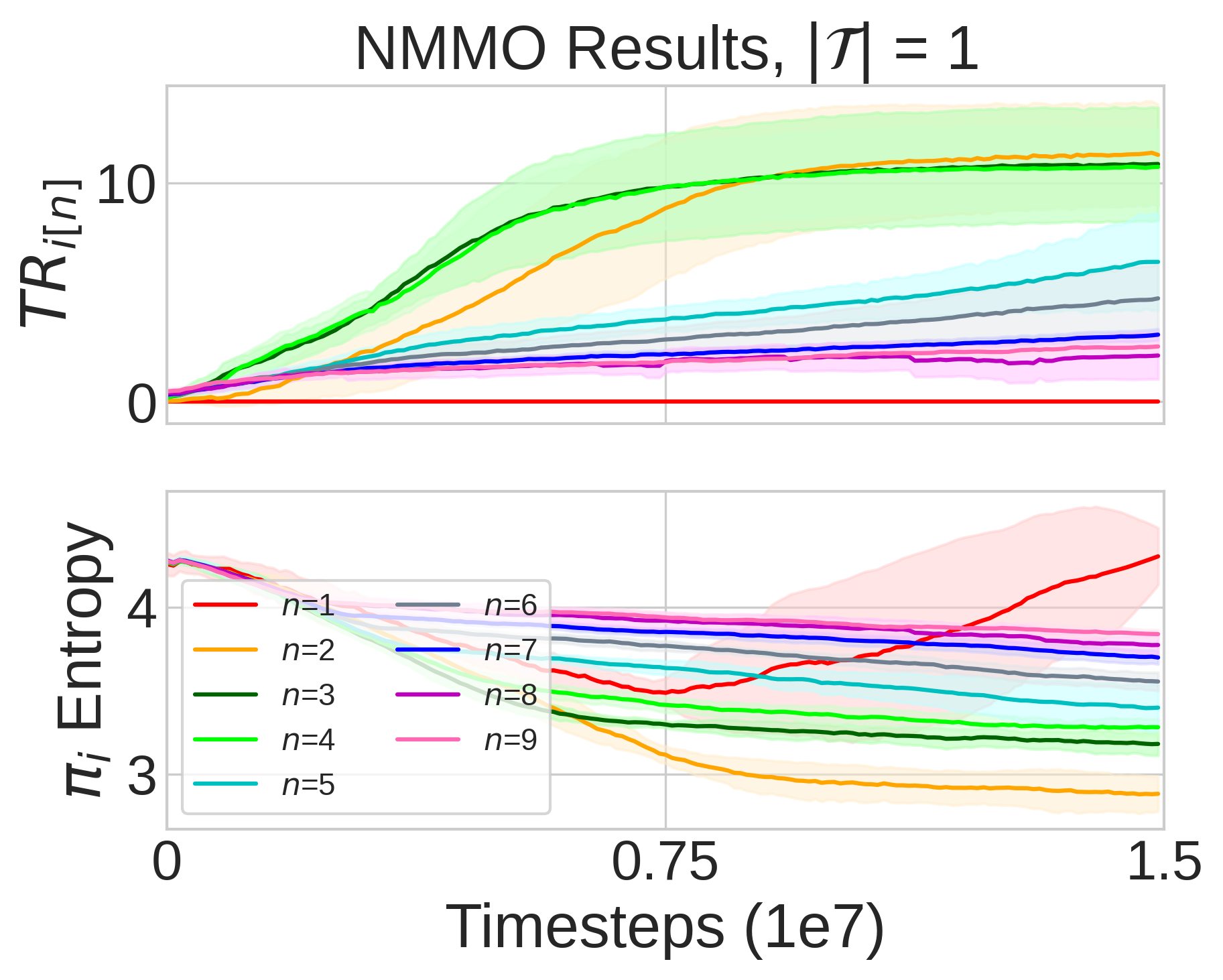}
        \caption{NMMO Results.}
        \label{fig:nmmo_results}
    \end{subfigure}
        \caption{(a) \textit{4-States:} Team reward (top) and mean difference in $Q$-values scaled by maximum $Q$-value (bottom). (b) and (c) \textit{Cleanup \& NMMO:} Team reward (top) and mean policy entropy (bottom). All plots show 95\% confidence intervals. Teams with few teammates (e.g., $n=2, 3, 4$) tend to perform best across all environments.}
\end{figure*}

\newtext{
\paragraph{Neural MMO.}
Neural MMO (NMMO)~\cite{suarez2019neural} is a large, customizable, and partially observable multiagent environment that supports foraging and exploration.
Agent observability is limited to an egocentric 15 $\times$ 15 pixel window and have movement and combat actions.
Agents maintain a stash of consumable resources (food and water) that deplete by 0.02 each timestep (minimum of 0) but are replenished by 0.1 by harvesting in the environment (maximum of 1.0 each).
Agents share rewards and resources to resemble a hunter-gatherer society.
There is no standard NMMO configuration; therefore, we reward agents for positive increases to their lowest resource: $\min(I)^{t} - \min(I)^{t-1}$ when $\min(I)^{t} > \min(I)^{t-1}$, where $I$ is the inventory of food and water.
Agents must learn to maintain both food and water to receive reward, creating multiple dynamically changing reward-causing state-action pairs, a more challenging scenario than the other environments.
We implement PPO agents for eight trials of $1.5 \times 10^7$ environmental timesteps (1,000 per-episode) using Rllib.
Additional details in Appendix H.1.
}

\section{Empirical Results}
\label{sec:results}



\newtext{
In this section, we evaluate how the size of teams affects team performance.
We observe a similar trend across all \camera{environments and learning algorithms}: performance initially increases with more teammates, but decreases once teams become too large.
Thus, our results highlight a ``sweet spot'' team structure that helps guide agents' learning in these environments.
}

\subsection{4-States Environment Results}
\label{sec:4-states_results}

Due to the small number of states, larger teams in 4-States can generate more reward, even if agents act randomly (more agents can collect a reward of 1 in $s_r$).
Thus, we measure team reward as a fraction of each team structure's theoretical optimal reward assuming no randomness (mean episode reward of $\frac{1}{2}$ for $n=1$, $\frac{n-1}{n}$ for $n>1$).
Figure~\ref{fig:4_state_results} (top) shows the team reward compared to optimal ($y$-axis) over timesteps of our experiments ($x$-axis).
Each line represents a different team size with 95\% confidence intervals.
When $n=1$, only $25.2\%$ of the optimal reward is achieved.
Increasing to $n=2$ dramatically increases the reward to $66.5\%$ of the optimal solution, and larger teams result in diminishing returns.
Considering $\epsilon$-exploration and transitions impose about 33\% randomness,
$n=2$ performs well.

The $y$-axis of Figure~\ref{fig:4_state_results} (bottom) shows the mean difference in $Q$-values between actions, scaled by the maximum $Q$-value in the table at each timestep.
Lower values indicate agents expect similar values for any action and have not learned the reward dynamics of the environment.
This plot follows the same trend as the reward: agent learn more disparate $Q$-values when $n=2$, but larger teams cause agents to learn similar values for all actions.
This indicates a decrease of environmental information as $n$ grows.

\newtext{
Figure~\ref{fig:spatial_4states} shows the team state visitation frequencies as a fraction of the optimal policy with 95\% confidence intervals (i.e., transitioning between $s_c$ to $s_r$ when $n=1$, and one agent in $s_c$ while $n-1$ agents in $s_r$ when $n > 1$).
When $n=1$, the agent fails to learn the value of transitioning to $s_c$.
Agents perform closest to optimal when $n=2$, suggesting they learn the value of visiting both $s_c$ and $s_r$ while avoiding $s_3$ and $s_4$.
The agents are unable to fully converge to the optimal policy due to the stochastic transition function and $\epsilon$-greedy action selection.
With larger $n$, agents tend to visit $s_c$ more often than optimal and $s_r$ with less frequency, suggesting they fail to learn the reward-causing dynamics of the environment with larger groups, supporting our theory.
}

\begin{figure}[t]
    \centering
    \includegraphics[width=0.95\linewidth]{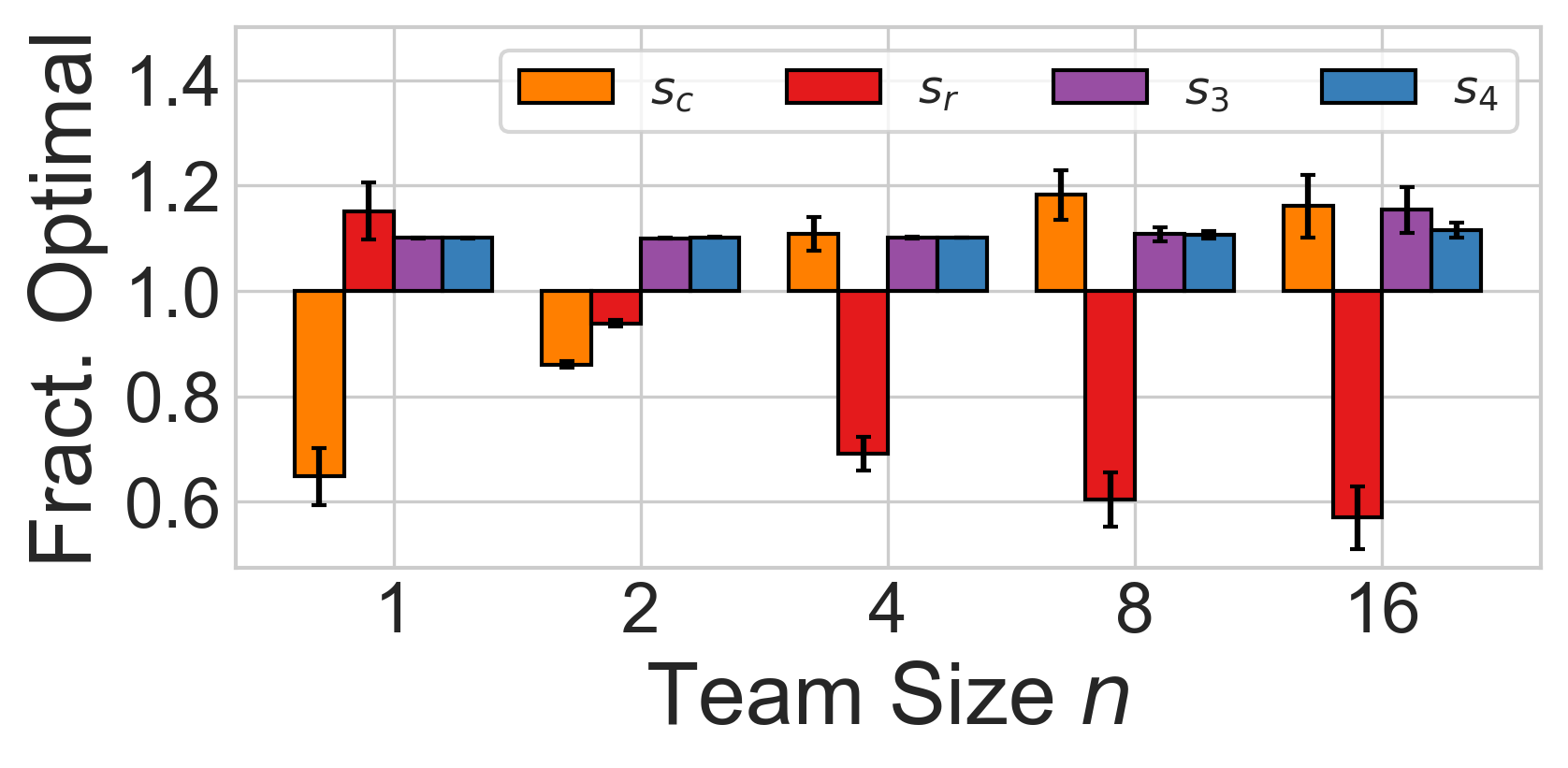}
    \caption{\textit{4-States:} Mean state visitation fraction of optimal joint policy for different team sizes (95\% confidence intervals). Positive bars indicate more visits than optimal and negative bars indicate fewer. Team size $n=2$ performs closest to the optimal joint policy.}
    \label{fig:spatial_4states}
\end{figure}

\subsection{Cleanup Gridworld Game Results}
\label{sec:cleanup_results}

\newtext{
Figure~\ref{fig:cleanup_results} shows the team reward (top) and mean policy entropy (bottom) along the $y$-axes with 95\% confidence intervals, and timesteps along the $x$-axis.
Our results follow a similar trend as seen in 4-States.}
\addressreview{
More reward is initially obtained by adding teammates and is highest when $n=2$ or $n=4$, due to a division of labor: half of the agents specialize in each role of cleaning the river or picking apples.
When $n=3$, two agents specialize in river-cleaning roles while only one collects apples, causing slightly less team reward due to more sharing than when $n=2$, but collecting fewer apples than when $n=4$.}
\newtext{
Team structures with $n > 4$ tend to have decreasing team reward, following our theoretical findings in Section~\ref{sec:decrease_state_info}.
}
\newtext{
We use \textbf{policy} entropy ($\pi_i$ entropy) to better understand role specialization on teams, where lower $\pi_i$ entropy implies higher role specialization and less random actions (Figure~\ref{fig:cleanup_results} bottom).
We observe that when $n=2$, mean $\pi_i$ entropy is lowest and as $n$ increases, agents policies tend to become more random.
We observe a correlation between team reward and agents' convergence to specialized roles, measured by $\pi_i$ entropy, and find the lowest mean $\pi_i$ entropy when $n=2$.
This $\pi_i$ entropy tends to increase as $n$ is larger, suggesting agents' policies become more random.
}

\newtext{
Figure~\ref{fig:spatial_cleanup} shows the mean team reward three agents receive (columns) at different map locations when they are in a team of $n=4$ (top row) and $n=6$ (bottom row), where darker red indicates more reward.
The plots for the remaining teammates in each row are shown in Figure 9
in Appendix G.2.
When $n=4$, we find that the two agents that specialize in river-cleaning roles (agent 0 and 3) also spatially divide the labor into different parts of the river, one in the top half and one in the bottom half.
This allows their two other teammates (agents 1 and 2) to collect apples and reward for the team.
However, when $n=6$ we observe that three agents specialize in river-cleaning roles (agents 0, 3, and 4), but are less specialized in their cleaning locations.
Agents 0 and 3 tend to clean the same segment of the river, converging to redundant policies that do not generate significantly more apples for their apple-picking teammates to collect.
}

\subsection{Neural MMO Results}
\label{sec:nmmo_results}

Figure~\ref{fig:nmmo_results} shows the NMMO results.
When $n=1$, the agent fails to learn the value of collecting both food and water which results in no reward.
As teammates are introduced, the agents learn complimentary harvesting roles and gain the highest team reward when $n=2,3,4$. However, we observe   diminishing returns with larger teams (when $n>4$).     
We hypothesize that these values are highly correlated with the number of inventory item types and harvesting tasks.
Similar to Cleanup, agents have significantly less $\pi_i$ entropy in these settings, suggesting that agents have converged to specific roles on their team and act less randomly than when they have no teammates or many teammates.
This result supports our theory and is consistent with our other experiments: a sufficient number of teammates results in more favorable policies, but too many teammates leads to diminishing returns.
Our spatial results in NMMO, similar to the findings in Cleanup in Figure
~\ref{fig:spatial_cleanup}, are shown in Figure 11
in Appendix H.2.




\begin{figure}[t]
    \centering
    \includegraphics[width=0.95\linewidth]{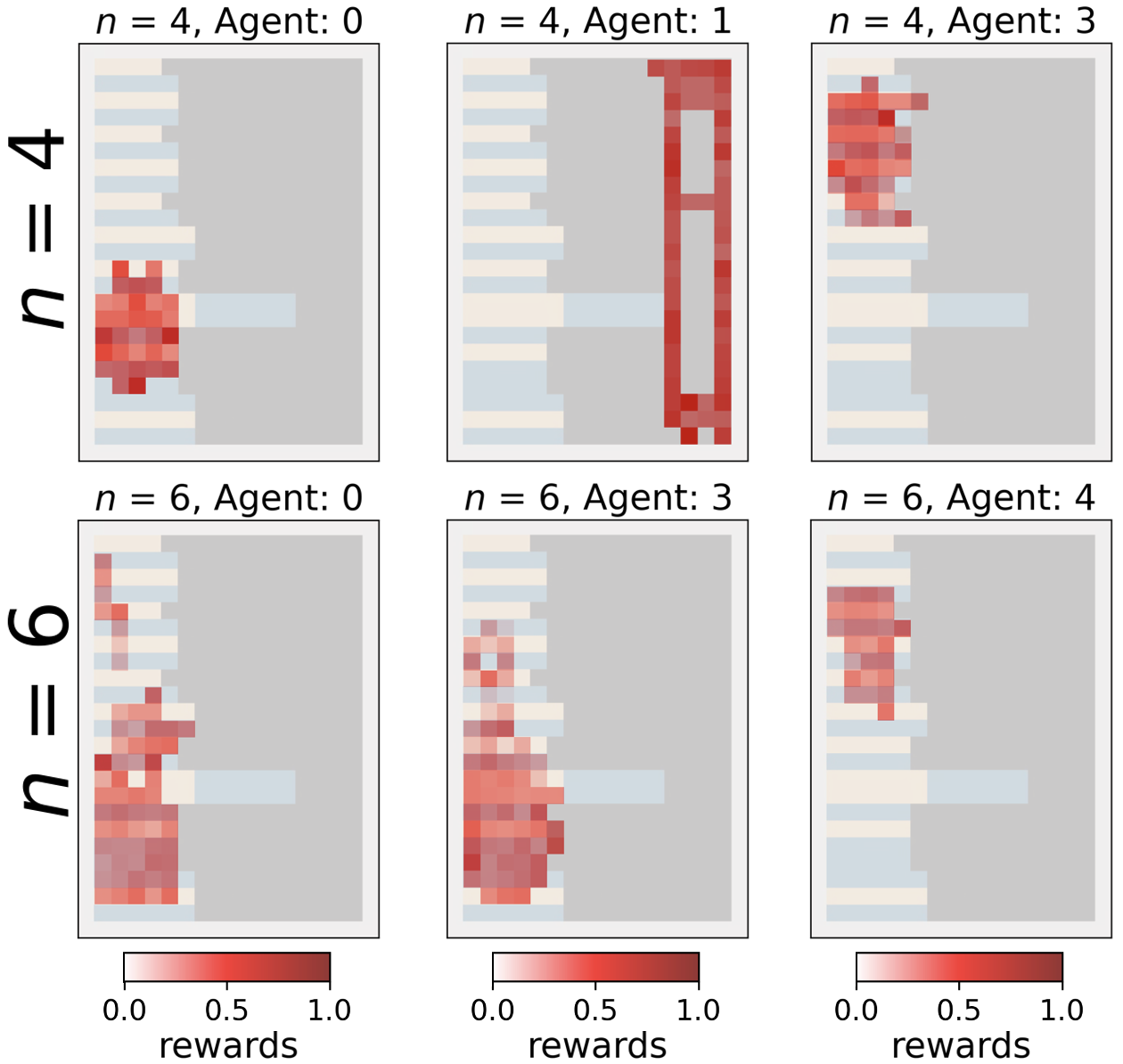}
    \caption{\textit{Cleanup:} Team reward obtained at each pixel for different agents when $n = 4$ (top row) and when $n = 6$ (bottom row). Agents learn specialized roles when $n=4$ but converge to overlapping redundant roles when $n=6$. Plots for all agents in $T_i$ for both $n=4$ and $n=6$ settings are shown in Appendix G, Figure 9.} 
    \label{fig:spatial_cleanup}
\end{figure}

\section{Discussion}
\label{sec:discussion}

Our research provides an understanding as to why, and under which conditions, smaller teams can outperform larger teams.
Introducing teammates can help agents identify reward-causing state-action pairs (Section~\ref{sec:teams_redistribute_reward}), but too many teammates can make credit assignment more difficult which hinders learning (Section~\ref{sec:negatives_team_size}).
This provides theoretical explanations behind the empirical results of several recent papers~\cite{Durugkar2020BalancingIP,Radke2022Exploring}.

\newtext{
A common perception about RL theory is that convergence to the optimal policy is guaranteed given infinite computation.
While this finding is true for single-agent RL~\cite{sutton2018reinforcement}, convergence guarantees are known to not hold in many multiagent settings.
Our paper's context of multiagent teams, even in a scenario with one team (i.e., cooperative population), is a setting where convergence to an optimal joint policy \textbf{is not} guaranteed, even with infinite computation.
This is a result of our finding in Theorem~\ref{thm:entropy_to_zero}, since information converges to zero as a function of team size.
However, information does not need to be zero for RL to fail (it can fail when information is sufficiently small); thus, in practice, infinitely large team size is not required for this result.
We are unable to guarantee non-convergence since random policy updates could \emph{theoretically} result in the optimal joint policy; however convergence to this policy is not guaranteed.
}


\newtext{
While we provide much needed insights into the importance of teams and team structures to shape the learning problems and reward functions for individual learning agents, there are several opportunities for future work.
These include developing social planning algorithms to construct highly efficient team structures within a population of agents from domain variables, precisely measuring $\epsilon$ and $\mu$ from domain variables, alternate definitions of teams or reward schemes, and developing agents that can regulate their own team alignment to overcome a sub-optimal team structure situation~\cite{radke2023tuning}.
Further progress in this direction will be pivotal to better understand how and in what social scenarios cooperation and complex behaviors can naturally emerge at individual and group levels.
}

\section*{Acknowledgements}
This research is funded by the Natural Sciences and Engineering Research Council of Canada (NSERC), an Ontario Graduate Scholarship, a Cheriton Scholarship, and the University of Waterloo President's Graduate Scholarship.
We thank the Vector Institute for providing the compute resources necessary for this research to be conducted.
We also thank Alexi Orchard, Valerie Platsko, Kanav Mehra, and all reviewers for their feedback and useful discussion on earlier drafts of this work.

\bibliographystyle{named}

\clearpage
\newpage
\appendix

\section{Two State Environment Stochastic Game}

The two state environment showed in Figure~\ref{fig:2_states} of the main text induces a stochastic game whenever $n>1$.
This stochastic game has multiple possible Nash Equilibria on which teammates must coordinate on.

\begin{figure}[h]
    \centering
    \includegraphics[width=\linewidth]{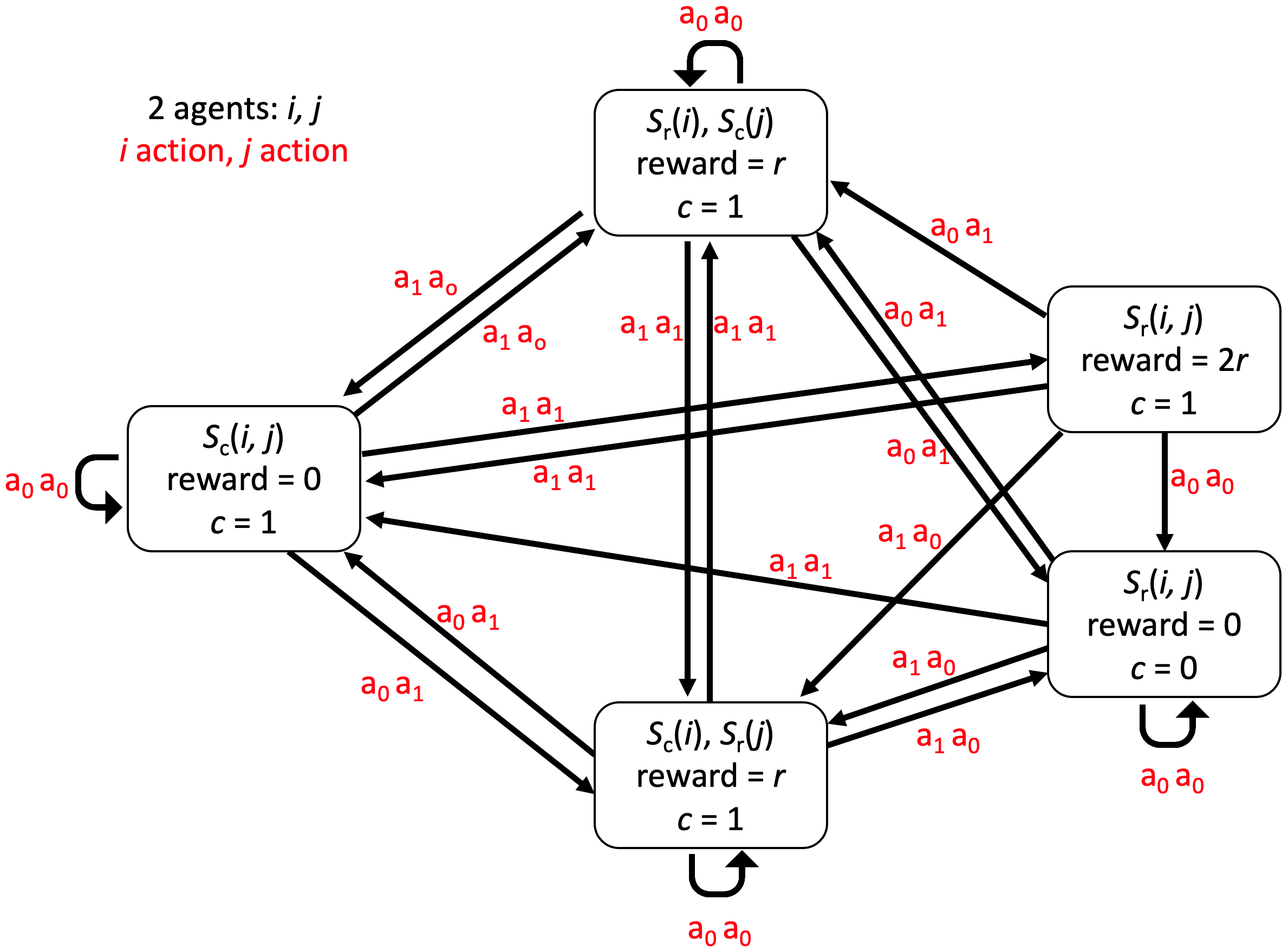}
    \caption{Stochastic game diagram induced from our two state environment. Game states are labeled so that $s_c(i,j)$ represents both agents ($i$ and $j$) being in physical state $s_c$.}
    \label{fig:sg}
\end{figure}

To show the emergence of multiple Nash Equilibria, Figure~\ref{fig:sg} shows the stochastic game induced in this environment with $n = 2$ agents ($i$ and $j$).
The possible scenarios of the game are labeled so that $s_c(i,j)$ represents both agents being in physical state $s_c$.
The reward represents the total reward yielded from the environment in that specific game state (i.e., reward = $2r$ represents both $i$ and $j$ received $r$).
For any agent to obtain the reward of $r$ at $s_r$, some agent in the environment must visit $s_c$ to change the boolean signal to $c = 1$.
With just two agents, there are multiple joint policies that yield optimal reward on which agents must learn to coordinate on.
Specifically, the two agents could 1) both move between $s_c$ and $s_r$ together, 2) transition from $s_c$ to $s_r$ (vice versa) with $a_1$ to never be in the same state, or 3) each agent always stays in either $s_c$ or $s_r$ using $a_0$.

\section{Reward Redistribution}

The first theoretical finding in the manuscript is how larger teams increase the probability of agent $i$ receiving a positive reward signal for executing a reward-causing state-action pair.

\begin{thm}
There exists an environment where increasing team size increases the probability of an agent receiving a reward for executing any reward-causing state-action pair that is greater than if they were not in a team.
\label{thm:apdx_reward_increase}
\end{thm}

\begin{proof}
Due to agent's policies being initialized uniformly at random at the beginning of learning, we assume full coverage of the state space by all independent agents in the limit.
Subsequently, suppose agent $i$ is executing a reward-causing state-action pair that yields the minimum reward in the environment (Assumption 3).
Any teammate moving to a reward state increases the reward $i$ receives for executing that reward-causing state-action pair through $TR_{i[n]}$ compared to when $i$ acts individually.
The probability of any teammate $j$ being in a reward state $s_r$ is equal to the product of agents \textbf{not} being in $s_r$ subtracted from 1.
Let $0 < \zeta < 1$ be the probability that a teammate $j$ is \textbf{not} located in a reward state, $s_r$, where $\zeta_j = \zeta_k$ for each $j, k \in T_i$ (i.e., $\zeta$ is assumed to be equal for all teammates).
For a team of size $n$, the probability of any teammate being in a reward state at any timestep is $P(s_j = s_r) = 1 - \zeta^{(n-1)}$.
Since $0 < \zeta < 1$, the second term $\zeta^{(n-1)} \rightarrow 0$ as $n \rightarrow \infty$.
As a result, the overall probability of any teammate being in a reward state $P(s_j = s_r)$ converges to $1$ as team size increases.







\end{proof}

Theorem~\ref{thm:apdx_reward_increase} shows how larger teams make reward-causing state-action pairs attractive for agents that learn from experience to maximize their future reward.

\section{Decreased Information}

Our second theoretical contribution examines the impact of team size on the amount of information agents gain through their policies.

\begin{proposition}
Let $\boldsymbol\pi_{T_i}$ be the joint fixed behavior policy of agents in $T_i$ that generates a joint trajectory of experiences $\boldsymbol\tau_{T_i}$
(a collection of individually observed trajectories by each $i \in T_i$), where the randomness of state-action pairs in $\boldsymbol\tau_{T_i}$ depends on all $N$ agents (by the definition of a stochastic game).
Let $TR_{i[n]}^t$ be a random variable denoting the team reward at any timestep $t$ (where the randomness of the deterministic reward follows from the randomness of the joint state-action pairs of individual agents in $T_i$ at time $t$, depending on all $N$ agents, $\boldsymbol\tau_{T_i}^{t}$).
It follows that:
\[ \mathcal{I}(Z(\boldsymbol\tau_{T_i}); \boldsymbol\tau_{T_i}^{t} | \boldsymbol\tau_{T_i}^{-t}) = \mathcal{H}(TR_{i[n]}^{t} | \boldsymbol\tau_{T_i}^{1:t-1}). \]
\label{prop:apdx_info_entropy}
\end{proposition}

\begin{proof}
The chain rule of mutual information gives us:

\begin{equation*}
    \begin{aligned}
    \mathcal{I}(Z(\boldsymbol\tau_{T_i}); \boldsymbol\tau_{T_i}^{t} | \boldsymbol\tau_{T_i}^{-t}) &= \mathcal{I}(Z(\boldsymbol\tau_{T_i}); \boldsymbol\tau_{T_i}^{t} , \boldsymbol\tau_{T_i}^{-t}) \\ &- \mathcal{I}(Z(\boldsymbol\tau_{T_i}); \boldsymbol\tau_{T_i}^{-t}) \\
    &= \mathcal{I}(Z(\boldsymbol\tau_{T_i}); \boldsymbol\tau_{T_i}) - \mathcal{I}(Z(\boldsymbol\tau_{T_i}); \boldsymbol\tau_{T_i}^{-t}).
    \end{aligned}
\end{equation*}

By the definition of mutual information, we can expand in terms of entropy:

\begin{equation*}
    \begin{aligned}
    = \mathcal{H}(Z(\boldsymbol\tau_{T_i})) &- \mathcal{H}(Z(\boldsymbol\tau_{T_i}) | \boldsymbol\tau_{T_i}) \\ &- \mathcal{H}(Z(\boldsymbol\tau_{T_i})) + \mathcal{H}(Z(\boldsymbol\tau_{T_i}) | \boldsymbol\tau_{T_i}^{-t})
    \end{aligned}
\end{equation*}

\[= \mathcal{H}(Z(\boldsymbol\tau_{T_i}) | \boldsymbol\tau_{T_i}^{-t}) - \mathcal{H}(Z(\boldsymbol\tau_{T_i}) | \boldsymbol\tau_{T_i}).\]


We know $Z(\boldsymbol\tau_{T_i})$ is a deterministic function of $\boldsymbol\tau_{T_i}$ due to the deterministic aggregation (mean reward) of $n$ deterministic reward functions of all teammates.
The deterministic individual reward functions are already dependent on all $N$ agents; thus, we can drop the second term and simplify to:

\[= \mathcal{H}(Z(\boldsymbol\tau_{T_i}) | \boldsymbol\tau_{T_i}^{-t}).\]


Since we know each agent in $T_i$ is optimizing their discounted sum of future team rewards, we know $Z(\boldsymbol\tau_{T_i}) = \Sigma_{t=1}^{H} \gamma^{t-1} TR_{i[n]}^{t}$, and can substitute for $Z(\boldsymbol\tau_{T_i})$:

\[ = \mathcal{H}(TR_{i[n]}^{t} | \boldsymbol\tau_{T_i}^{-t}) \]
\[= \mathcal{H}(TR_{i[n]}^{t} | \boldsymbol\tau_{T_i}^{1:t-1}, \boldsymbol\tau^{t+1:H}).\]


Finally, since $TR_{i[n]}^{t}$ is unable to be impacted by the future (i.e., anything greater than $t$), we can remove the correlation with $\boldsymbol\tau^{t+1:H}$:

\[= \mathcal{H}(TR_{i[n]}^{t} | \boldsymbol\tau_{T_i}^{1:t-1}).\]



\end{proof}

Proposition~\ref{prop:apdx_info_entropy} equates the information at any time of a stochastic game to the entropy of the team reward signal.
The left-hand side quantifies the information between a single joint state-action pair for the team $\boldsymbol\tau_{T_i}^{t}$ and the team's joint policy return over the joint trajectory, $Z(\boldsymbol\tau_{T_i})$, conditioned on the joint trajectory without timestep $t$, $\boldsymbol\tau_{T_i}^{-t}$.
Next, we show that the variance of the team reward function converges to zero as team size increases.

\begin{lem}
The team reward random variable $TR_{i[n]}$ for any state-action pair converges to the mean environmental reward (mean of any agent's individual reward function) as team size increases in the limit (i.e., $TR_{i[n]}(\vb{s}^t, \vb{a}^t, \vb{s}^{t+1}) \rightarrow \overline{R_i}$ as $n \rightarrow \infty$).
\label{lem:apdx_reward_distribution_exp}
\end{lem}

\begin{proof}
Since the team reward is an aggregation of $n$ individual and uniformly random rewards samples from identical reward functions, $TR_{i[n]} \approx \mathcal{N} \left( \overline{R_i}, \frac{\sigma^2_{R_i}}{\sqrt{n}} \right)$ by the Central Limit Theorem, where var$\left[ R_i \right] = \sigma_{R_i}^2$.
The variance var$\left[ TR_{i[n]} \right] = \frac{\sigma^2_{R_i}}{\sqrt{n}}$, with a derivative of var$\left[ TR_{i[n]} \right]' = - \frac{\sigma_{R_i}}{ \sqrt{n^{3}}}$.
Since $\sigma_{R_i} = \sqrt{\sigma^2_{R_i}}$ is the standard deviation of $R_i$ (i.e., distance from $\overline{R_i}$), we know $\sigma_{R_i} > 0$.
Furthermore, $\sigma_{R_i}$ is a constant and $n \geq 1$; thus, var$\left[ TR_{i[n]} \right]'$ is negative and converges to zero as $n$ increases in the denominator.
\end{proof}

Finally, we use Proposition~\ref{prop:apdx_info_entropy} and Lemma~\ref{lem:apdx_reward_distribution_exp} to show that the information in a stochastic game converges to zero as a funciton of team size.

\begin{thm}
The information in a stochastic game at time $t$, $\mathcal{I}(Z(\boldsymbol\tau_{i}); \boldsymbol\tau_{i}^{t} | \boldsymbol\tau_{i}^{-t})$, converges to 0 as the size of a team, $n$, increases in the limit.
\label{thm:apdx_entropy_to_zero}
\end{thm}

\begin{proof}
By Proposition~\ref{prop:info_entropy}, we can use the entropy of $TR_{i[n]}^t$ to determine the information of $Z(\boldsymbol\tau_{i})$ at time $t$ of a trajectory.
By the Central Limit Theorem and Lemma~\ref{lem:reward_distribution_exp}, let $TR_{i[n]}^t$ be a Gaussian distributed random variable so that $TR_{i[n]}^t \approx \mathcal{N} \left( \overline{R_i}, \frac{\sigma^2_{R_i}}{\sqrt{n}} \right)$.
For readability, let the variance $\sigma^2 = \frac{\sigma^2_{R_i}}{\sqrt{n}}$.
We rewrite the entropy of $TR_{i[n]}$ at time $t$ given the trajectory up to $t$, $\mathcal{H}(TR_{i[n]}^{t} | \boldsymbol\tau^{i}_{1:t-1})$, in terms of the function's variance:

\begin{align*}
    &\begin{aligned}
        \mathcal{H}(TR_{i[n]}^{t} | \boldsymbol\tau^{i}_{1:t-1}) &= - \int_{TR_{i[n]}} p(TR_{i[n]}) \log p(TR_{i[n]}) \\
        &= - \mathbb{E} \left[\log \mathcal{N}(\overline{R_i}, \sigma^2)\right] \\
        &= - \mathbb{E} \left[ \log \left[ \frac{1}{\sqrt{2\pi} \sigma^2 } e^{-\frac{1}{2} ( \frac{R_i - \overline{R_i}}{\sigma^2})^2} \right] \right] \\
        &= \frac{1}{2} \log (2\pi \sigma^2) + \frac{1}{2\sigma^2} \mathbb{E} \left[ (R_i - \overline{R_i})^2 \right] \\
        &= \frac{1}{2} \log (2\pi \sigma^2) + \frac{1}{2}
    \end{aligned}
\end{align*}

Since $\pi$ is a constant, the variance $\sigma^2 = \frac{\sigma^2_{R_i}}{\sqrt{n}}$ regulates the entropy of $TR_{i[n]}^t$.
By Lemma~\ref{lem:reward_distribution_exp}, we know $\Lim{n\to\infty} \frac{\sigma^2_{R_i}}{\sqrt{n}} \rightarrow 0$.
Thus, the entropy and information carried by the actions of $\pi_i$ in a stochastic game at time $t$ converges to zero as team size increases.






\end{proof}

Theorem~\ref{thm:apdx_entropy_to_zero} states that agents will be unable to perform proper credit assignment and learn good policies as their team's size increases in the limit.
This result is significant since it characterizes how fully cooperative systems can perform worse than a population of multiple smaller teams.

\section{Information with Teams}

A fixed behavior policy $\pi_i$ induces a stationary visitation distribution for agent $i$ over states and state-action pairs, denoted as $d^{\pi_i}(s)$ and $d^{\pi_i}(s,a)$ respectively.
Since we are concerned with the progression of how agents learn, our theory assumes agents are initialized with random policies that cover the state space uniformly, consistent with past work~\cite{arumugam2021information}.

The value of var$\left[ \mathcal{I}^{\pi_{i}}_{S_i,A_i} (Z_{T_i}) \right]$ depends on calculating the KL Divergence for state-action pairs from the distribution of states and actions for $\pi_i$, $d^{\pi_i}$.
Given the distributional support $\mathcal{X}_{s_i, a_i}$ (the distribution of team rewards conditioned on specific state-action pairs that are not mapped to zero), this can be expanded to be:

\begin{align*}
    &\textrm{var}\left[ \mathcal{I}^{\pi_{i}}_{S_i,A_i} (Z_{T_i}) \right] = \\ & \textrm{var}_{s_i, a_i \sim d^{\pi_i}} \left[ \sum_{Z_{T_i} \in \mathcal{X}_{s_i, a_i}} p(Z_{T_i}|s_i,a_i) \log \left( \frac{p(Z_{T_i}|s_i,a_i)}{p(Z_{T_i}|s_i)} \right) \right]
    \label{eq:kl_expanded}
\end{align*}

Note that $S_i$ and $A_i$ are based on agent $i$'s individual observations and policy, but $Z_{T_i}$ is based on their shared team reward.

\section{4-States}

\begin{figure}[t]
    \centering
    \includegraphics[width=0.8\linewidth]{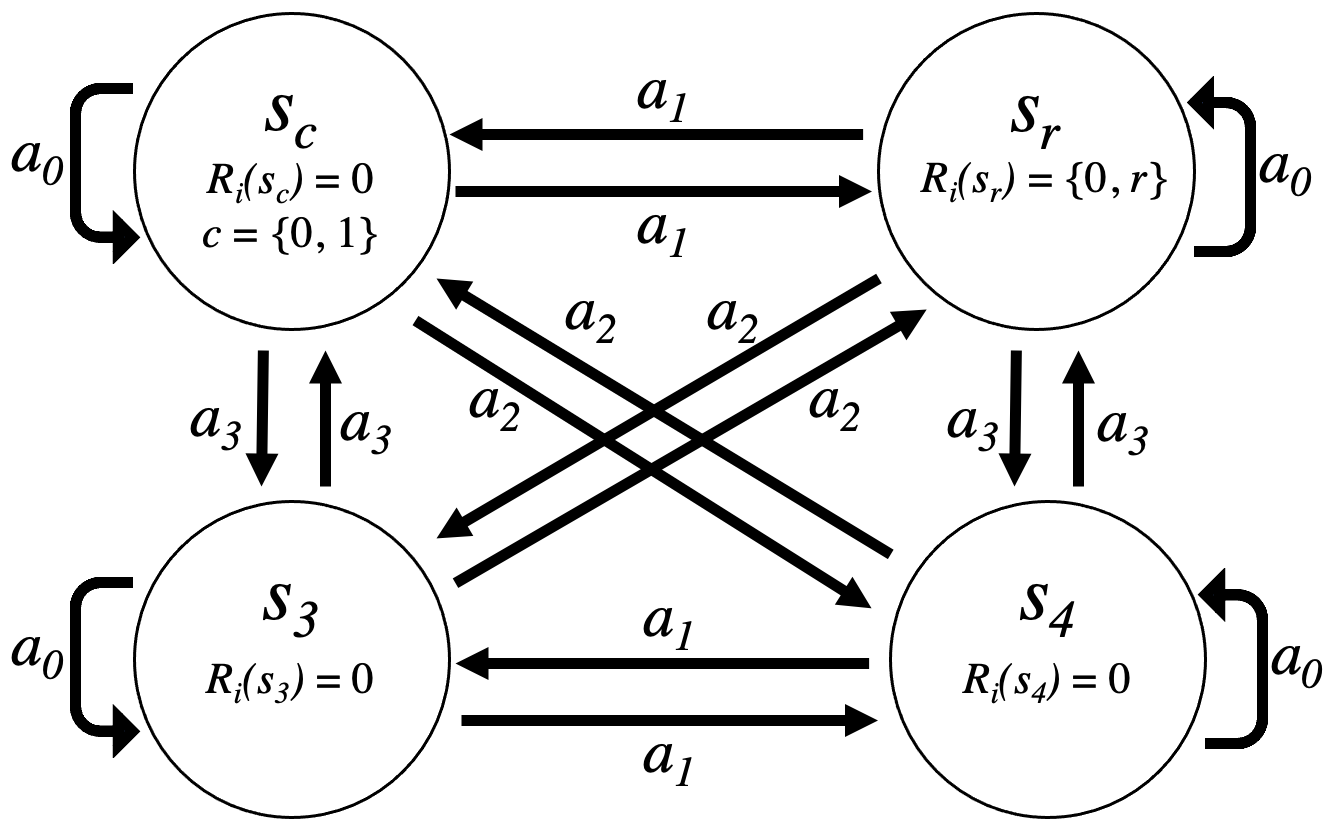}
    \caption{\textit{4-States:} Environment diagram.}
    \label{fig:4states_diagram}
\end{figure}

\subsection{Environment}

\newtext{
Figure~\ref{fig:4states_diagram} shows the 4-States environment in our evaluation, an augmentation of the simple 2-States environment shown in Figure~\ref{fig:2_states} of the main text.
We add two ``no-op'' states to the two state environment that return no reward and do not impact the binary signal (i.e., agents should avoid these states).
States are labeled $s_c$ , $s_r$, $s_3$ (no-op), and $s_4$ (no-op).
A reward of +1 is given at $s_r$, conditioned on the visitation of $s_c$.
Agents simultaneously choose among four actions: stay at their current state ($s_0$) or move to any of the other three states ($s_1$, $s_2$, or $s_3$).
An action transitions agents to their intended next state with 90\% probability and to another random state with $10\%$ probability.
We fix $|\mathcal{T}| = 1$ and increase $n$ by a factor of 2 to remove the impact of other teams on the binary signal.
Agents using Tabular $Q$-Learning~\cite{sutton2018reinforcement} with $\gamma = 0.9$ and $\epsilon$-exploration ($\epsilon=0.3$) for 50 trials of 1,000 episodes (100 steps each).
The stochastic transitions and $\epsilon$-exploration causes agents not to select the best action or move to their intended state about 33\% of timesteps.
}


\section{Iterated Prisoner's Dilemma (IPD)}

\subsection{Environment}

We follow a similar IPD configuration as recent work with teams~\cite{Radke2022Exploring,radke2022importance} and assume that there is a cost ($c$) and a benefit ($b$) to cooperating where $b>c>0$.
Agents are randomly paired with another agent at each timestep, a \emph{counterpart}, that may or may not be a teammate with some probability $\nu$.
Agents must choose to either cooperate with ($C$) or defect on ($D$) their counterpart.
Agents only observe the team label (i.e., number) of their counterpart, and receive their team reward, $TR_{i[n]}$, after their own and teammates' interactions; therefore, the strategies of all agents on team $T_i$ affects how agents learn to play any member of $T_i$.
We fix the cost $c = 1$, benefit $b = 5$, and define $|\mathcal{T}| = 2$ with increasing sizes of each team where $n=1$ (no teams), $n=2$ (one teammate), and then multiples of 5 to study general trends with larger teams.
We fix $\nu = 97\%$ (non-teammates are 16 times more likely than teammates) and 100\% when $n = 1$ (agents do not play themselves).
Each experiment lasts $1.0 \times 10^6$ episodes where $N=30$ agents learn using Deep $Q$-Learning~\cite{Mnih2015HumanlevelCT}, repeated for 20 trials.

\subsection{Results}

\begin{figure}[t]
    \centering
    \includegraphics[width=0.8\linewidth]{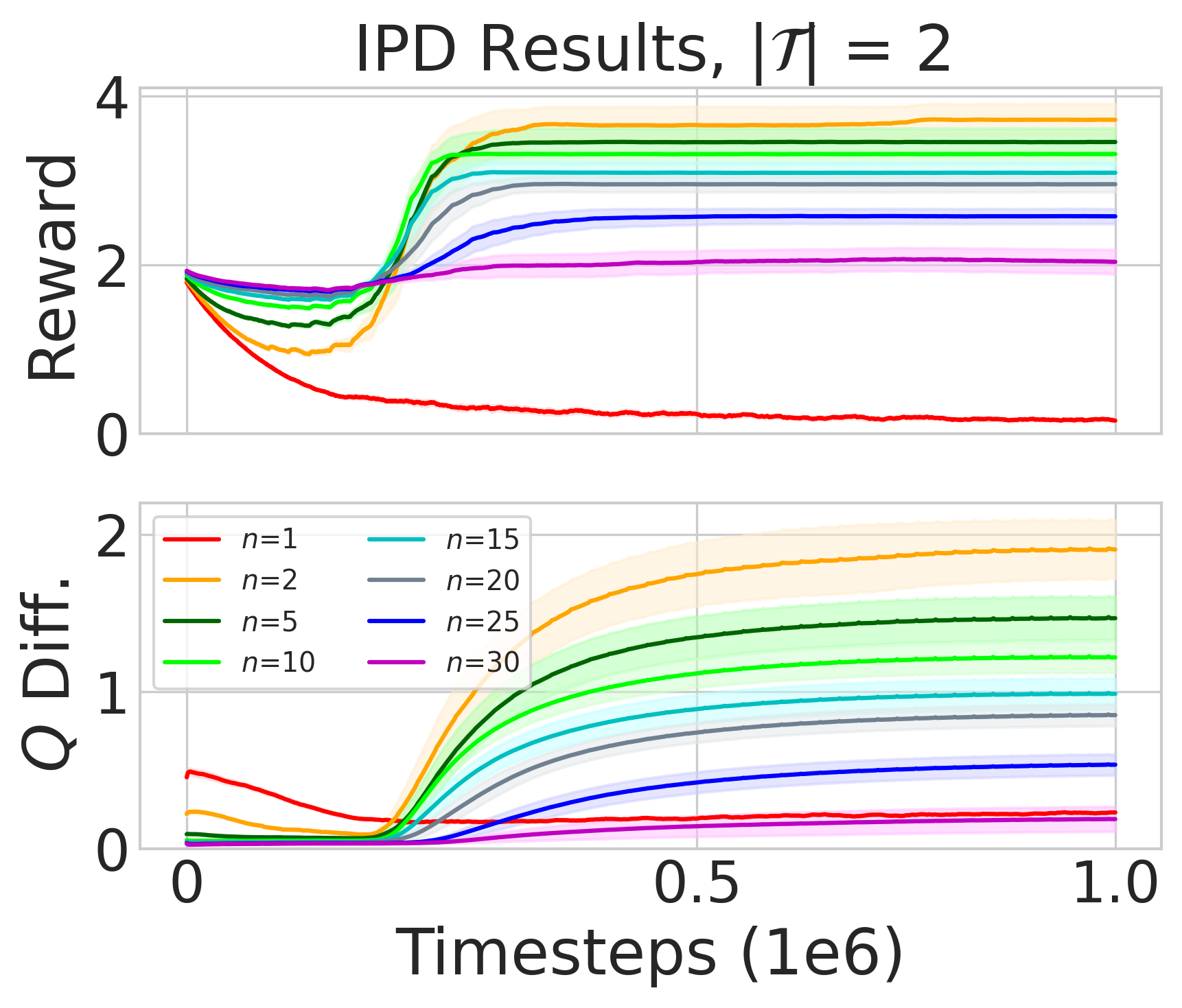}
    \caption{\textit{IPD:} Mean population reward (top) and mean difference in agents' $Q$-values (bottom). Less difference between $Q$-values indicates agents have less preference for either action.}
    \label{fig:ipd_results}
\end{figure}

\begin{figure}[t]
    \centering
    \includegraphics[width=0.8\linewidth]{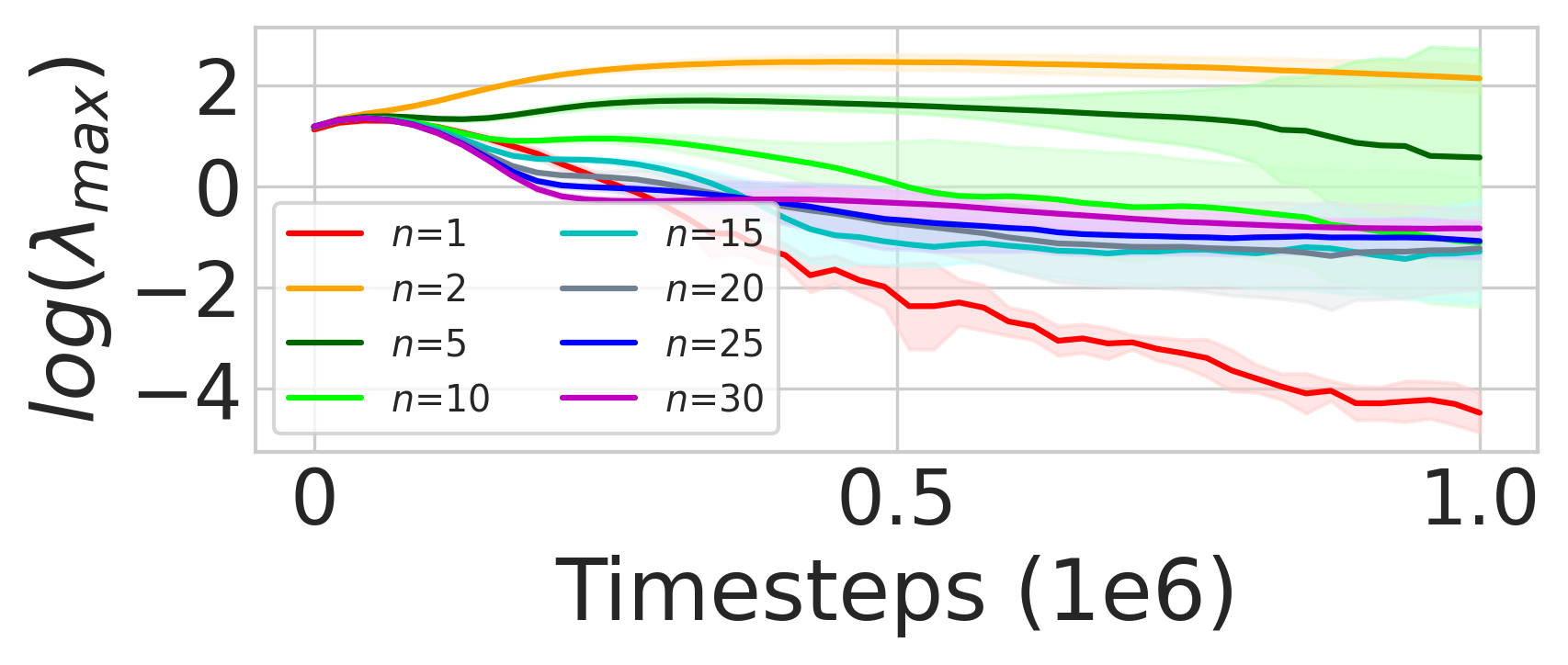}
    \caption{\textit{IPD:} Mean maximum eigenvalue ($\lambda_{max}$) of agents' Hessian matrices (i.e., flatness of loss landscape).}
    \label{fig:eig_max}
\end{figure}

\begin{figure*}[t]
    \centering
    \includegraphics[width=0.85\linewidth]{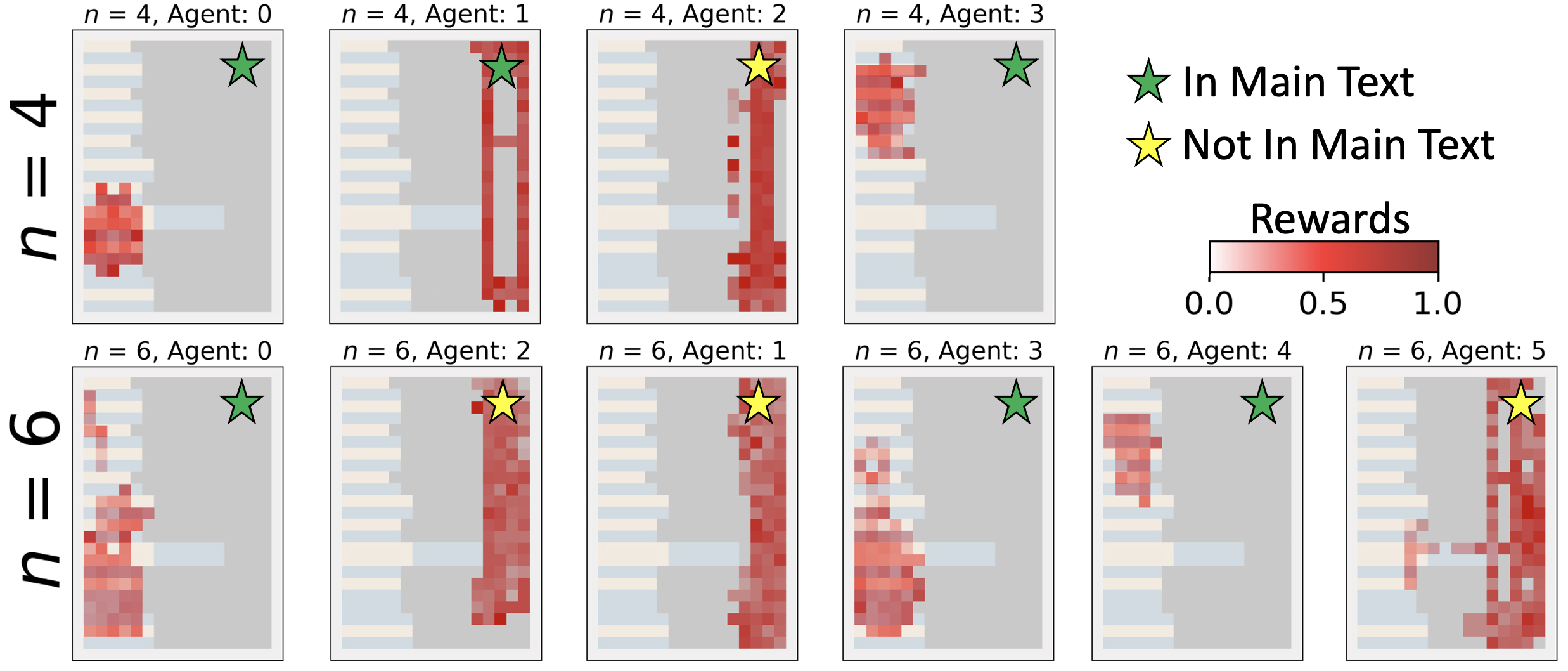}
    \caption{\textit{Cleanup:} Team reward obtained at each pixel for different agents. The top row shows all agents' behaviors when $n=4$ and the bottom row shows all agents when $n=6$. Plots that appear in the main text are indicated with a green star and plots that are omitted from the main text due to space limitations are indicated with a yellow star.}
    \label{fig:cleanup_spatial_full}
\end{figure*}

\begin{figure}[t]
    \centering
    \includegraphics[width=0.7\linewidth]{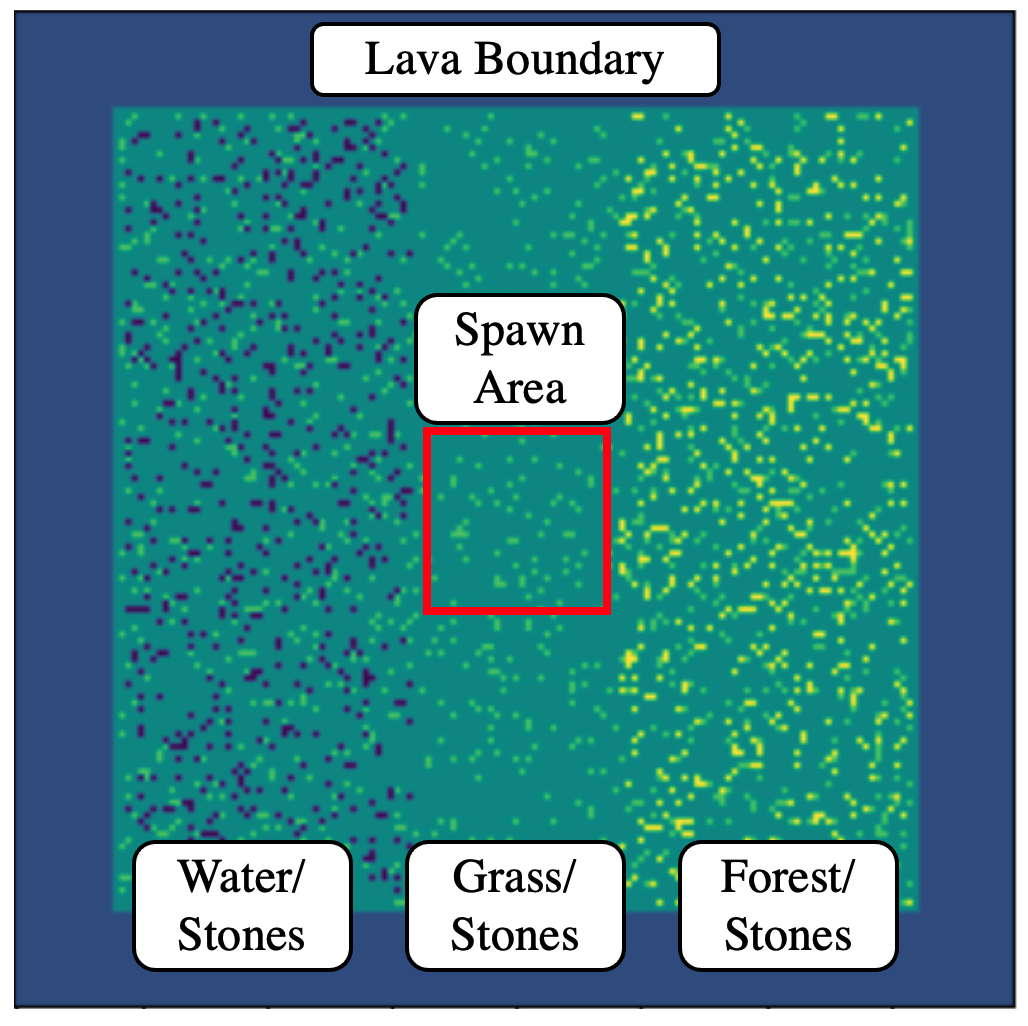}
    \caption{\textit{NMMO:} Custom environment layout for our evaluation.}
    \label{fig:nmmo_layout}
\end{figure}

Figure~\ref{fig:ipd_results} shows our results in the IPD environment for the mean population reward (top) and the difference in $Q$-values for $C$ and $D$ when paired with non-teammates (bottom).
Both graphs share the same $x$-axis, representing the timesteps of our experiments.

Since mutual cooperation is the result with the highest mean population reward, we use reward as a proxy for learned cooperation (higher is better).
When $n=1$, agents converge to the Nash Equilibrium of mutual defection and obtain the lowest mean population reward.
Consistent with past work~\cite{Radke2022Exploring}, our results show how having even one teammate allows agents learn cooperation and achieve high mean population reward despite only being paired with this teammate 3\% of the time.
However, team growth has diminishing returns.
When $n=30$, the mean population reward approaches the mean reward and agents behave randomly (i.e., $\overline{R_i} = 2$ when cost is 1, benefit is 5).

The bottom graph shows how initially providing agents with teammates ($n=2$) increases the difference in $Q$-values significantly since agents learn the benefit of mutual cooperation.
Agents adapt this behavior towards other teams and the population experiences high cooperation and high reward.
Further increasing team size tends to reduce the difference in $Q$-values until agents have little $Q$-value difference when $n=30$.
These results are consistent with our theory and experiments in the other three domains.

As a further analysis into how teams impact learning, Figure~\ref{fig:eig_max} shows the mean maximum eigenvalue ($\lambda_{max}$) of agents' policy network Hessian matrices as they learn (log$_{10}$ scale).
Lower values of $\lambda_{max}$ represent a flatter optimization surface~\cite{kaur2022maximum} that makes convergence through stochastic gradient descent more difficult.
When $n=1$, the high rate of 0 reward leads to a flat optimization landscape, but when $n=2$ or $5$, $\lambda_{max}$ is the highest among all team structures we study.
As teams grow larger, the loss landscape flattens and convergence to a minima becomes more difficult.
This highlights that teams shape the loss landscape to assist convergence to a cooperative minima~\cite{Radke2022Exploring}, but large team structures flatten the landscape and reduce convergence.

\section{Cleanup Gridworld Game Extended}

\subsection{Environment}
\label{sec:appendix_cleanup_env}

Cleanup~\cite{SSDOpenSource} is a temporally and spatially extended Markov game representing a sequential social dilemma.
Agents in Cleanup have eight actions: 9 movement (up, down, left, right, stay, turn left, and turn right), a cleaning beam, and a punishment beam.
Agent observability is limited to an egocentric 15 $\times$ 15 pixel window, and
agents receive +1 reward for collecting an apple in the orchard.
Apple growth is conditional on the cleanliness of an adjacent river, and cleaning this river yields no direct environmental reward.
Successful groups in Cleanup balance the temptation to free-ride and pick apples with the public obligation to clean the river.
We set $|\mathcal{T}|=1$ and increase team size to remove impacts of other teams on the conditional reward structure.
We implement Proximal Policy Optimization (PPO)~\cite{PPO2017} agents for 10 trials of $1.6 \times 10^8$ episodes (1,000 timesteps each) using the Rllib RL library.

\subsection{Spatial Results}

Figure~\ref{fig:cleanup_spatial_full} shows the spatial behavior of all agents in one trial when $n=4$ (top row) and $n=6$ (bottom row).
This figure is an expanded version of Figure~\ref{fig:spatial_cleanup} in the main text, where darker red corresponds with higher reward when the agent is located at that spatial location.
When $n=4$ (top row), the population divides labor so that Agents 0 and 3 agents specialize to clean the river and Agents 1 and 2 pick apples which achieves the highest reward in our evaluation, shown in Figure~\ref{fig:cleanup_results} of the main text.
Additionally, Figure~\ref{fig:cleanup_spatial_full} (top row) shows how Agents 0 and 3 not only both converge to clean the river, but learn different cleaning \emph{roles} and spatially divide the river territory for more efficiency.
This spatial specialization is not typically observed with apple picking agents, but both apple picker agents still collect a significant amount of apples when $n=4$ regardless.

\begin{figure*}[t]
    \centering
    \includegraphics[width=\linewidth]{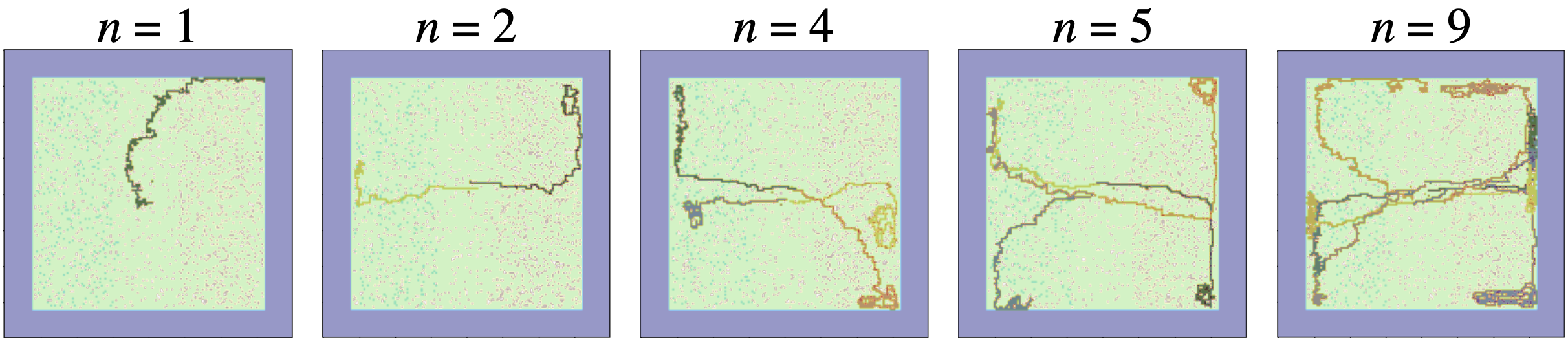}
    \caption{\textit{NMMO:} Agent behavior in NMMO when $n=1,2,4,5,9$. When $n=1$, the agent spends time in the center region of the map which results in no reward. Agents learn about the value of food and water when they have teammates. When $n=2$ or $n=4$, agents spatially disperse and specialize in roles of collecting food or water while not interfering with each other. When $n=5$ or $n=9$, agents begin to converge to similar areas of the map and eventually interfere with each other's ability to collect food or water and venture back into the center area of the map.}
    \label{fig:nmmo_spatial}
\end{figure*}

Compare this with when $n = 6$ shown in Figure~\ref{fig:cleanup_spatial_full} (bottom row), where we consistently find 3 river cleaner and 3 apple picker policies emerge within agents in $T_i$.
The behavior of the three river cleaners is less spatially specialized, resulting in Agents 0 and 3 cleaning the same location and learning the same role on their team (i.e., the role of cleaning the bottom half of the river).
This duplication of roles leads to less team reward than smaller team structures despite having more agents, as shown in Figure~\ref{fig:cleanup_results} of the main text.
Since we observe that two spatially specialized agents are able to effectively clean the river (seen when $n=4$), the team would benefit from one of these redundant cleaners learning to instead pick apples and collect more reward.
This gives further insight into why large cooperative systems achieve less reward than systems composed of multiple smaller teams in Cleanup, even when mixed incentives exist between teams as shown in~\cite{Radke2022Exploring}.

\section{Neural MMO Extended}

\subsection{Environment}

\newtext{
Neural MMO (NMMO)~\cite{suarez2019neural} is a large, customizable, and partially observable multiagent environment that supports foraging and exploration.
We configure a map with 1024 $\times$ 1024 pixels bounded by lava tiles to enclose the agents within the environment.
As mentioned in the main text, agent observability is limited to an egocentric 15 $\times$ 15 pixel window and have movement and combat actions.
Agents maintain a stash of consumable resources (food and water) that deplete some amount at each environmental timestep but are replenished through harvesting from the lakes and forests located throughout the environment.
There is no standard NMMO configuration; therefore, we can customize the environment and reward function to satisfy the assumptions made in Section 4 (shown in Figure~\ref{fig:nmmo_layout}).
Agents in a team share water and food resources amongst themselves and we remove agent death by starvation so that every episode is the same length.
Agents always spawn in a random location at the center of the map.
The environment has stones which agents must move around to reach water and forest tiles.
Grass tiles offer nothing to the agents.
}

\newtext{
We set a resource depletion rate of -0.02 (minimum of 0.0), replenish amount of +0.1 (maximum amount of 1.0), and spatially separate the forests and lakes to encourage exploration.
We reward agents for positive increases to their lowest resource: $\min(I)^{t} - \min(I)^{t-1}$ when $\min(I)^{t} > \min(I)^{t-1}$, where $I$ is the inventory of food and water.
Agents must learn to maintain both food and water to receive reward, creating multiple dynamically changing reward-causing state-action pairs, a more challenging scenario than the other environments.
We implement PPO agents for 5 trials of $1.6 \times 10^7$ episodes (1,000 timesteps each) using Rllib.
}

\subsection{Spatial Results}

\newtext{
Figure~\ref{fig:nmmo_spatial} shows the movement of agents when $n=1,2,4,5,9$.
When $n=1$ (Figure~\ref{fig:nmmo_spatial} left), the agent has difficulty learning about the value of both food and water, resulting in the agent staying in the center region of the map where there is only grass and stone (Figure~\ref{fig:nmmo_layout}).
When the agent is given a teammate ($n=2$; Figure~\ref{fig:nmmo_spatial} middle left), they converge to complimentary roles and explore different regions of the environment, collecting either food or water and sharing their resources.
This behavior is also observed when $n=4$ with two agents collecting food or water each.
This joint policy generates one of the best team reward results in our evaluation showing the benefits of adding teammates.
When $n=5$ or $n=9$, the agents still learn complimentary roles; however, they tend to interfere with each other and cover similar areas of the environment, consistent with our spatial results in Cleanup shown in Figure~\ref{fig:spatial_cleanup} of the main text or Figure~\ref{fig:cleanup_spatial_full} in Appendix G.2.
The environment is significantly large so that this movement is avoidable; however, agents have difficulty learning how to spatially disperse as to maximize the reward from their joint policy.
Furthermore, when $n=9$, two agents return to the center grass/stone area later in an episode which contributes no positive reward for their team.
}

\section{Summary of Notation}

Table~\ref{tab:notation} lists the notation used throughout the paper for easy access for the reader.

\begin{table*}[h]
\begin{center}
\begin{tabular}{ |p{2.2cm}|p{12.5cm}| }
\hline
Notation & Description  \\
\hline\hline
    $i$ & An arbitrary agent. \\
    $j$ & A second arbitrary agent. \\
    $\mathcal{N}$ & Set of all agents. \\
    $N$ & Size of the set of all agents. \\
    $A$ & Joint action space. \\
    $S$ & Joint state space. \\
    $R$ & Joint reward space. \\
    $P$ & Transition function. \\
    $\gamma$ & Discount factor. \\
    $\Sigma$ & Policy space of all agents. \\
    $\pi_i$ & Policy of agent $i$. \\
    $t$ & Arbitrary timestep of an episode. \\
    $s_i$ & Single state for agent $i$. \\
    $a_i$ & single action for agent $i$. \\
    $\vb{s}^t$ & Joint state at time $t$. \\
    $\vb{a}^t$ & Joint action at time $t$. \\
    $R_{i}^t(\vb{s}^t,\vb{a}^t,\vb{s}^{t+1})$ & Agent $i$'s individual reward at time $t$. \\
    $V_i$ & Value function of agent $i$. \\
    $\mathcal{T}$ & Set of all teams. \\
    $\mathcal{T}_i$ & Set of teams $i$ belongs to. \\
    $T_i \in \mathcal{T}_i$ & Specific team that $i$ belongs to. \\
    $n$ & The number of agents in a team. \\
    $TR_{i[n]}$ & Team reward for a team of size $n$. \\
    $H$ & Length of a full episode. \\
    $\tau_i$ & Trajectory of state-action pairs generated by $i$. \\
    $\pi_{T_i}$ & Joint policy for $n$ agents in team $T_i$. \\
    $\tau_{T_i}$ & Joint trajectory for $n$ agents in team $T_i$. \\
    $\tau^t_{T_i}$ & Joint state-action pair at time $t$ for the agents in team $T_i$. \\
    $\tau_{T_i}^{1:t-1}$ & Joint trajectory for $n$ agents in team $T_i$ up to time $t-1$. \\
    $\tau_{T_i}^{-t}$ & Joint trajectory for $n$ agents in team $T_i$ without the joint state-action pair at time $t$. \\
    $Z(\tau_{T_i})$ & Random variable denoting the team random return obtained from a joint trajectory $\tau_{T_i}$. \\
    $\vb{s}_{T_i}$ & Team $T_i$'s joint state. \\
    $\vb{s}_{T_i}$ & Team $T_i$'s joint action. \\
    $Z_{T_i}$ & Random variable denoting the team reward observed at $\vb{s}_{T_i}$ and taking joint action $\vb{a}_{T_i}$. \\
    $\mathcal{I}^{\pi_{i}}_{s_i,a_i}$ & Information gained by $\pi_i$ in single-agent setting. \\
    $D_{KL}$ & Kullback-Leibler (KL) divergence. \\
    $p(Z_i|s_i,a_i)$ & Distribution of returns conditioned on particular state-action pair. \\
    $p(Z_i|s_i)$ & Distribution of returns conditioned only on state. \\
    $\mathcal{I}(A_i;Z_i|S_i)$ & Expected information $\pi_i$ carries in single-agent setting. \\
    $\mathcal{I}^{\pi_{i}} (A_i;Z_{T_i}|S_i)$ & Expected information $\pi_i$ carries in a multiagent team from a team reward. \\
    $\mathcal{I}^{\pi_{i}}_{S_i,A_i} (Z_{T_i})$ & Expected information gained by $\pi_i$ over distribution of individual state-action pairs. \\
    $\epsilon$ & Threshold on the expected information in an environment. \\
    $\mu$ & Threshold on the variance of expected information across state-action pairs. \\
    $\mathcal{H}(TR_{i[n]}^{t})$ & Entropy of team reward funciton. \\
\hline
\end{tabular}
\end{center}
\caption{Notation summary throughout the paper for the reader.}
\label{tab:notation}
\end{table*}

\end{document}